%% file: main.tex
\documentclass[11pt]{article}

\usepackage[final]{acl}

\usepackage{times}
\usepackage{latexsym}

\usepackage[T1]{fontenc}

\usepackage[utf8]{inputenc}

\usepackage{microtype}

\usepackage{inconsolata}

\usepackage{graphicx}
\usepackage{amsmath}
\usepackage[dvipsnames,table]{xcolor}
\usepackage{color,soul}
\usepackage{subcaption}
\usepackage{hyperref}
\usepackage{amsfonts}
\usepackage{amssymb}
\usepackage{mathtools}
\usepackage[english]{babel}
\usepackage{booktabs}
\usepackage{multirow}
\usepackage{makecell}
\usepackage{bm}
\usepackage{fancyvrb}
\usepackage{enumitem}

\definecolor{PineGreen}{HTML}{008B72}
\definecolor{Cherry}{HTML}{D2042D}

\title{Toward Better Temporal Structures for Geopolitical Events Forecasting}

\author{
 \textbf{Kian Ahrabian\thanks{~{Authors contributed equally.}}} \And
 \textbf{Eric Boxer}\footnotemark[1] \And
 \textbf{Jay Pujara} \AND
 {\normalfont University of Southern California, Los Angeles, USA}\\
 Information Sciences Institute, Marina del Rey, USA\\
 \texttt{\{ahrabian,eboxer\}@usc.edu}, \texttt{jpujara@isi.edu} \\
}

\begin{document}
\maketitle

\input{sections/0.abstract}
\input{sections/1.introduction}
\input{sections/2.related}
\input{sections/3.preliminaries}
\input{sections/4.htkgh}
\input{sections/5.dataset}
\input{sections/6.experiments}
\input{sections/7.conclusion}

\section*{Limitations}
\paragraph{Domain}
Throughout this work, we focused on the geopolitical events due to their popularity; however, there are a plethora of other domains that could use such formalization using HTKGH, which we hope to see in future work.

\paragraph{Graph Neural Networks}
In this work, we focused on LLMs due to their flexibility and ease of use out of the box, but it is possible to investigate advanced existing and novel architectures based on the GNNs' framework, which might also solve the bottleneck issues we faced.
While \autoref{sec:exp} and \autoref{app:gnn} present a small study on a few GNN-based models, an in-depth and comprehensive experimentation with these models is warranted, which we leave to future work.

\paragraph{Link Prediction}
Given the aforementioned nuances regarding metrics and prediction strategies, there is a need for further investigations into the link prediction task.
While we present a limited set of experiments for the impact of HTKGH on the link prediction task in \autoref{sec:exp}, we leave further explorations to future work.

\paragraph{Knowledge Representation}
One of the often overlooked criteria for graph structures is the knowledge representation perspective, which concerns itself with querying time and complexity using technologies such as SPARQL.
Ideally, we want structures that both benefit learning algorithms and querying technology, demanding studies similar to the work by~\citet{10.1007/978-3-031-47240-4_15} for static knowledge graphs.
Given the limited scope of this work, we leave such investigations to future work.

\section*{Acknowledgment}
This material is based upon work supported by the Defense Advanced Research Projects Agency (DARPA) under Agreements No. HR00112590089 and No. HR00112220046.

\bibliography{custom}

\appendix

\input{sections/99.appendix}

\end{document}

%% file: sections/0.abstract.tex
\begin{abstract}
Forecasting on geopolitical temporal knowledge graphs (TKGs) through the lens of large language models (LLMs) has recently gained traction.
While TKGs and their generalization, hyper-relational temporal knowledge graphs (HTKGs), offer a straightforward structure to represent simple temporal relationships, they lack the expressive power to convey complex facts efficiently.
One of the critical limitations of HTKGs is a lack of support for more than two primary entities in temporal facts, which commonly occur in real-world events.
To address this limitation, in this work, we study a generalization of HTKGs, \textit{Hyper-Relational Temporal Knowledge Generalized Hypergraphs} (HTKGHs).
We first derive a formalization for HTKGHs, demonstrating their backward compatibility while supporting two complex types of facts commonly found in geopolitical incidents.
Then, utilizing this formalization, we introduce the \texttt{htkgh-polecat} dataset, built upon the global event database POLECAT.
Finally, we benchmark and analyze popular LLMs on our dataset, providing insights into 1) the positive impact of utilizing the HTKGH formalization compared to existing ones and 2) LLMs' adaptability and capabilities in complex forecasting tasks\footnote{\href{here}{~https://github.com/usc-isi-i2/htkgh-polecat}}.
\end{abstract}

%% file: sections/1.introduction.tex
\section{Introduction}

\begin{figure}
    \centering
    \includegraphics[width=\linewidth]{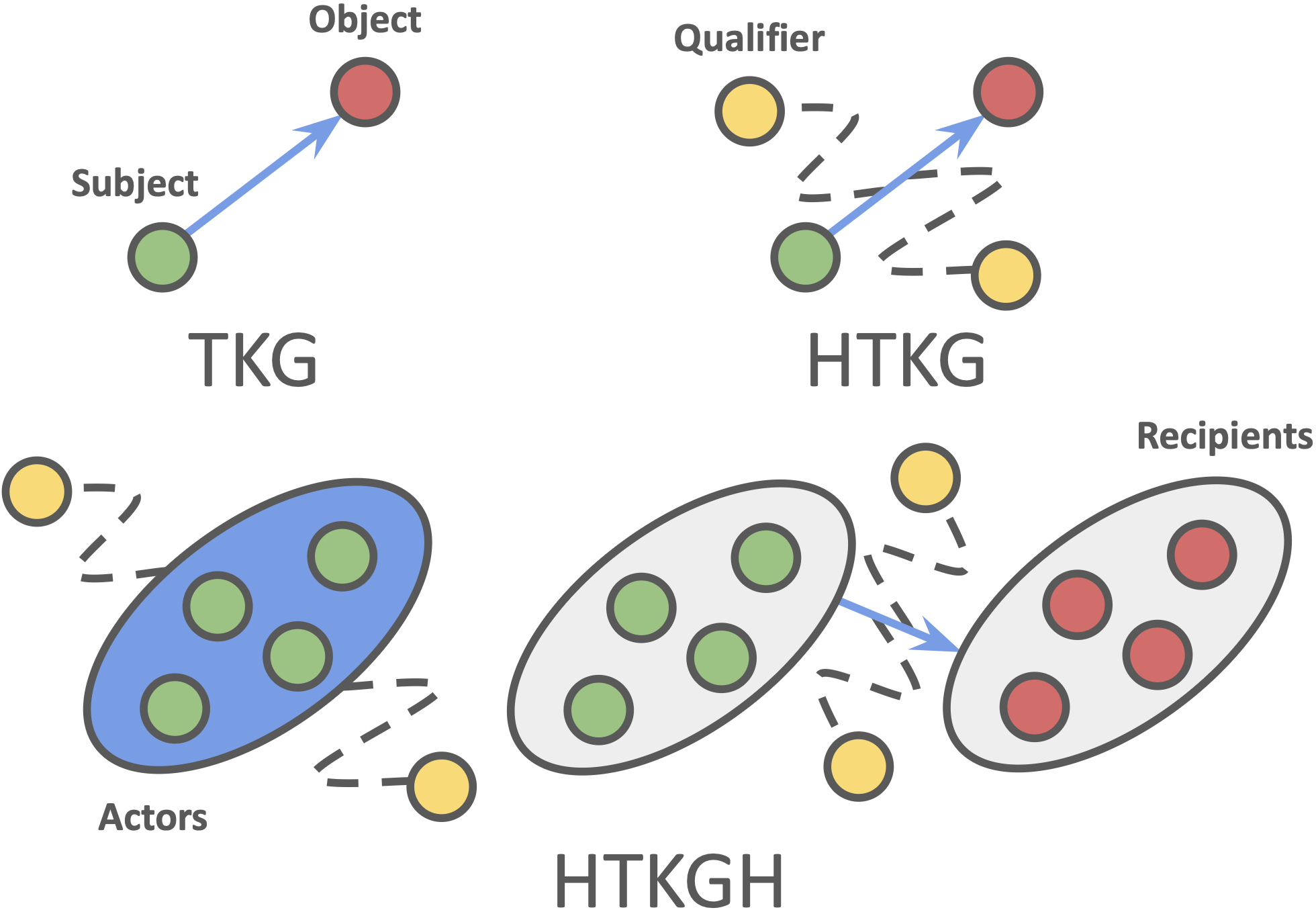}
    \caption{An overview of how facts are expressed in TKGs, HTKGs, and HTKGHs. HTKGHs add efficient support for complex facts involving many (\textit{i.e.,} more than two) primary entities, expanding the possible expressions to scenarios such as multi-national treaties.}
\label{fig:motiv}
\end{figure}

Temporal knowledge graphs (TKGs) are one of the most common structures used to express temporal relational facts.
In its simplest form, a TKG stores information as quadruples, each consisting of two entities, a relation, and a timestamp, which express a typed, directed temporal edge.
Recent extensions of TKGs, such as hyper-relational temporal knowledge graphs (HTKGs), allow additional information to be added to the edge in the form of qualifiers, pairs of relations and entities~\citep{ding-etal-2024-temporal}.
Apart from storing information, these structures are commonly used to run prediction queries, finding the most likely element(s) of a given partial fact.
One of the more challenging types of such queries is forecasting, which requires the model to make predictions in a time range it has not seen (see \autoref{sec:prel}).
Making predictions in the geopolitical domain is one of the recurrent use cases of such forecasting queries~\citep{garcia-duran-etal-2018-learning,jin-etal-2020-recurrent,10.5555/3737916.3742366}.
However, a common occurrence in geopolitical events is the involvement of numerous primary entities in an event, which creates a complex fact that is outside the norms of the two-primary-entity system that HTKG relies on.
In this work, we aim to address this issue through a generalization of the HTKG structure (see \autoref{fig:motiv} for an overview).

Examining the stored events in databases such as POLECAT~\citep{halterman2023plover}, we observe that many events involve group(s) of entities interacting with each other.
Specifically, two types of such events are more common: 1) \textbf{Group-Type:} a group of entities (>2) having the same mutual relationship, and 2) \textbf{Set2Set-Type:} a group of entities taking an action against another group of entities (>2 total entities).
However, trying to express such events within existing structures is difficult, as the simple structure of directed edges between two entities fails to properly represent them without imposing further issues, such as sparsity (see \autoref{sec:htkg_lim}).
To mitigate this problem, this work derives a formalization for hyper-relational temporal knowledge generalized hypergraphs (HTKGHs), which support an arbitrary number of primary entities and second-order edges, enabling efficient expression of more complex geopolitical events (see \autoref{sec:htkgh}).
Moreover, we introduce a new dataset, \texttt{htkgh-polecat}, which utilizes the HTKGH formalization, based on the POLECAT database.
Our statistics show that roughly one in four facts in \texttt{htkgh-polecat} follows one of the newly supported complex formats (see \autoref{sec:data}).

In recent years, LLMs have been extensively studied as a new approach to solving TKG queries, as they offer an easy-to-use approach that works out of the box with many structures via small amounts of prompting~\citep{lee-etal-2023-temporal,xia-etal-2024-chain}.
Such capabilities mainly stem from the extensive pre-training stage they undergo, where they are exposed to copious amounts of information, learning complex syntactical and semantic relationships between concepts and entities.
Notably, these capabilities extend beyond simple memorization, as they have demonstrated some level of general pattern recognition even when both semantic and syntactic information are hidden~\citep{lee-etal-2023-temporal}.
In this work, we evaluate the reasoning capabilities of LLMs over complex facts using the newly introduced dataset.
Our experiments highlight the impressive adaptiveness of LLMs when reasoning over information that is hidden or tampered with in a way that goes against the models' prior beliefs (see \autoref{sec:exp}).
Moreover, we demonstrate the empirical performance and inference time improvements that can be achieved using HTKGH over the existing structures.

To summarize, our contributions are as follows:
\begin{itemize}[itemsep=0.01cm]
    \item We highlight the shortcomings of HTKG and present an analysis of HTKGH as a generalization that enables efficient expression of complex facts common in geopolitical events.
    \item We introduce \texttt{htkgh-polecat} and its variations, a new dataset based on POLECAT that utilizes HTKGH formalization and employs strict heuristics to ensure validity and density.
    \item We adapt and analyze popular LLMs and graph-based models on our dataset, showcasing critical gains that can be achieved using the HTKGH formalization on performance and inference time, while also highlighting the impact of various confounding factors.
\end{itemize}

%% file: sections/2.related.tex
\section{Related Work}

\paragraph{TKG Extensions}
TKGs are known to struggle with higher-order relational details~\citep{lu2025two}.
Within the TKG structure, reification~\citep{noy2006defining} represents n-ary facts with an intermediate node connected to several binary predicates.
\citet{chebba2018attributed} proposes attributed relations and modeling n-ary relations with blank nodes.
However, these approaches unnecessarily increase the graph size, leading to extreme inefficiencies.
Beyond the TKG structure, HTKGs~\citep{ding-etal-2024-temporal} allow contextual attributes by adding qualifiers, disambiguating between otherwise identical TKG quadruples.
N-TKGs~\citep{hou-etal-2023-temporal} extend TKGs' quadruples into n-tuples, where relations connect $n$ entities with assigned roles, allowing for n-ary facts.
UniHR~\citep{liu2024unihr} introduces a hierarchical data representation module (HiDR) to model heterogeneous information jointly.
However, these extensions fail to move beyond the two-primary-entity system.

\paragraph{TKG Forecasting}
Traditional methods for TKG forecasting include time-aware embedding models~\citep{goel2020diachronic,han2020xerte}, temporal random walk-based approaches~\citep{sun2021timetraveler,liu2022tlogic}, and graph-based approaches~\citep{zhu2021learning,li2021temporal,xu2023temporal}.
More recently, a wave of LLM-based approaches has emerged, leveraging the pattern recognition capabilities of LLMs.
\citet{lee-etal-2023-temporal} studies LLMs' ability to perform TKG forecasting based on in-context learning.
GenTKG~\citep{liao2023gentkg} incorporates retrieval-augmented generation (RAG) into their method.
zrLLM~\citep{ding2023zrllm} relies on LLMs' semantic reasoning abilities to model unseen relations.
Nonetheless, LLMs' reasoning capabilities on complex facts remain understudied.

\paragraph{Event Databases}
Geopolitical event databases are rich sources for building TKGs, offering chronologically ordered facts involving several actors in one or more actions (\textit{e.g.,} wars, treaties, etc.).
Previously, snapshots of GDELT~\citep{leetaru2013gdelt} and ICEWS~\citep{DVN/28075_2015}, both part of broad initiatives to catalog global human behavior and predict conflicts, were used as TKG reasoning benchmarks.
Other examples are ACLED~\citep{raleigh2010introducing,semnani-etal-2025-lemonade}, containing location and battle event information related to eight conflict countries in West and Central Africa from 1960 to the present, and UCDP~\citep{sundberg2012introducing,sundberg2013introducing}, containing global data on non-state conflicts from 1989 to 2024.
In this work, we utilize POLECAT~\citep{halterman2023plover,halterman2023creating}, which stores cooperative and hostile geopolitical events from 2018 to 2024.

\begin{figure*}[t]
    \centering
    \begin{subfigure}{0.28\linewidth}
        \includegraphics[width=\linewidth]{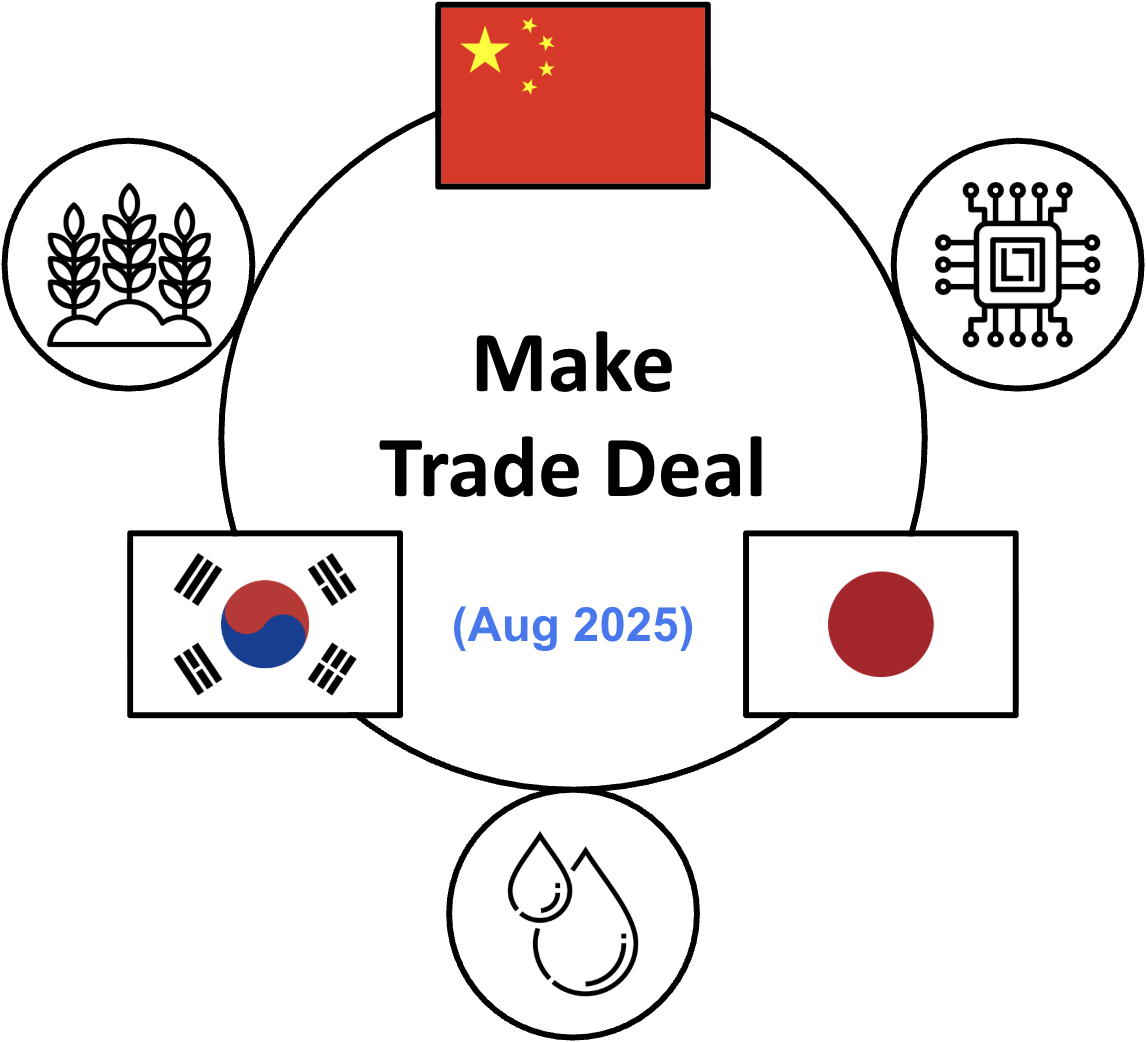}
        \caption{Original Fact}
    \end{subfigure}
    \hspace*{\fill}
    \begin{subfigure}{0.67\linewidth}
        \includegraphics[width=\linewidth]{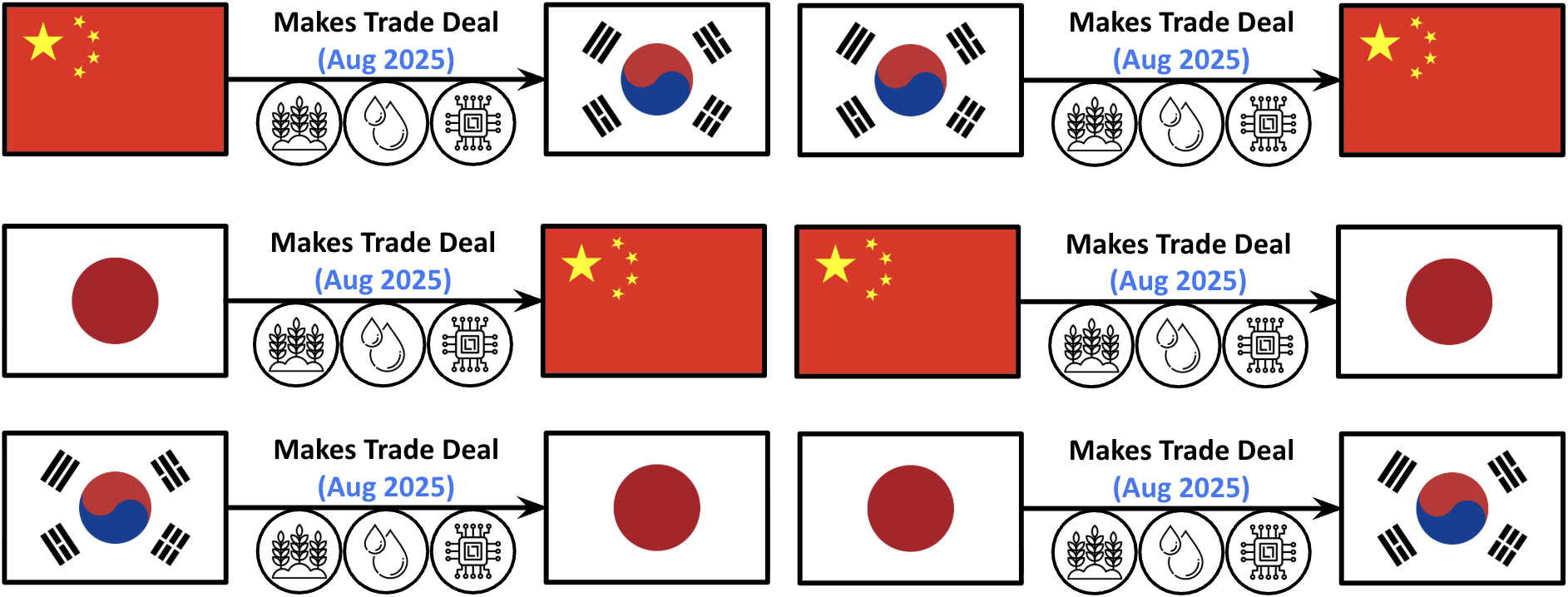}
        \caption{Decomposed Facts}
    \end{subfigure}
    \caption{\textbf{(a)} The original temporal fact representing a joint trade deal between $\{\text{China, Japan, and South Korea}\}$ on $\{\text{Cars, Chips, and Oil}\}$. \textbf{(b)} The decomposed version of the original fact designed for HTKGs. Notably, the decomposed version requires a lot of redundancy to represent. Moreover, without an additional entity representing the group of countries, it is indistinguishable from three separate trade deals between these countries.}
    \label{fig:orig_vs_decomp}
\end{figure*}

%% file: sections/3.preliminaries.tex
\section{Preliminaries}
\label{sec:prel}

\subsection{Definitions}
\paragraph{Temporal Knowledge Graph (TKG)}
Let $\mathcal{E}$, $\mathcal{R}$, and $\mathcal{T}$ be a set of entities, relations, and timestamps, respectively.
A TKG $\mathcal{G}$ comprises a set of quadruples representing temporal facts.
Formally, we define $\mathcal{G}$ as
\begin{equation}
    \mathcal{G} = \{ (s, r, o, t) \} \subseteq \mathcal{E} \times \mathcal{R} \times \mathcal{E} \times \mathcal{T} \  .
\end{equation}

\paragraph{Hyper-Relational Temporal Knowledge Graph (HTKG)}
Following the definition by \citet{ding-etal-2024-temporal}, we define an HTKG $\mathcal{H}$ as an attributed TKG where each quadruple is accompanied by a set of tuples containing additional information about the temporal fact.
Formally, let $\mathcal{G}$ be a TKG containing the \textit{primary quadruples}, we define $\mathcal{H}$ as
\begin{align}
    \mathcal{H} = \{ (f, Q) \; | \; f \in \mathcal{G}, Q \subseteq \mathcal{R} \times \mathcal{E} \} \  .
\end{align}
Since HTKGs can also represent any TKG, we only utilize them from this point forward.

\subsection{Queries}
\label{sec:queries}
\paragraph{Link Prediction} The most common query on HTKGs, which aims to find a missing entity to complete the partial fact at a particular time.
Formally, a link prediction query is defined as
\begin{equation}
    ((s, r, ?, t), Q) \;\;\; \text{or} \;\;\; ((?, r, o, t), Q) \  .
\end{equation}

\paragraph{Relation Prediction} The second most common query on HTKGs, which aims to find the missing relation in the partial fact at a particular time.
Formally, a link prediction query is defined as
\begin{equation}
    ((s, ?, o, t), Q) \  .
\end{equation}
While other queries, such as qualifier or time prediction, are possible, since they are not as commonly used, we will not discuss them in this work.

\subsection{Completion vs. Forecasting}
\label{sec:prelim:vs}
Orthogonal to the above query distinctions, each query can be either a 1) completion or 2) forecasting query based on the extent of access to historical information.
In a completion query, the predictive model has access to all temporal facts (\textit{i.e.}, the whole HTKG), before, concurrent, or after the query's timestamp.
However, in a forecasting query, the predictive model only has access to temporal facts before the query's timestamp.
Generally, forecasting queries are far more helpful in real-world decision-making scenarios, and this temporal constraint on information makes them much more challenging.
Throughout this work, we focus on the forecasting tasks.
Note that, a predictive model $\mathcal{M}$ is a forecaster capable of running forecasting queries if and only if it satisfies two constraints: 1) if the model has a training stage, all training datapoints must have a timestamp strictly smaller than all test samples, and 2) during the testing stage, the model does not have access to datapoints with equal or greater timestamp than the test datapoint.

\begin{figure}
    \centering
    \includegraphics[width=\linewidth]{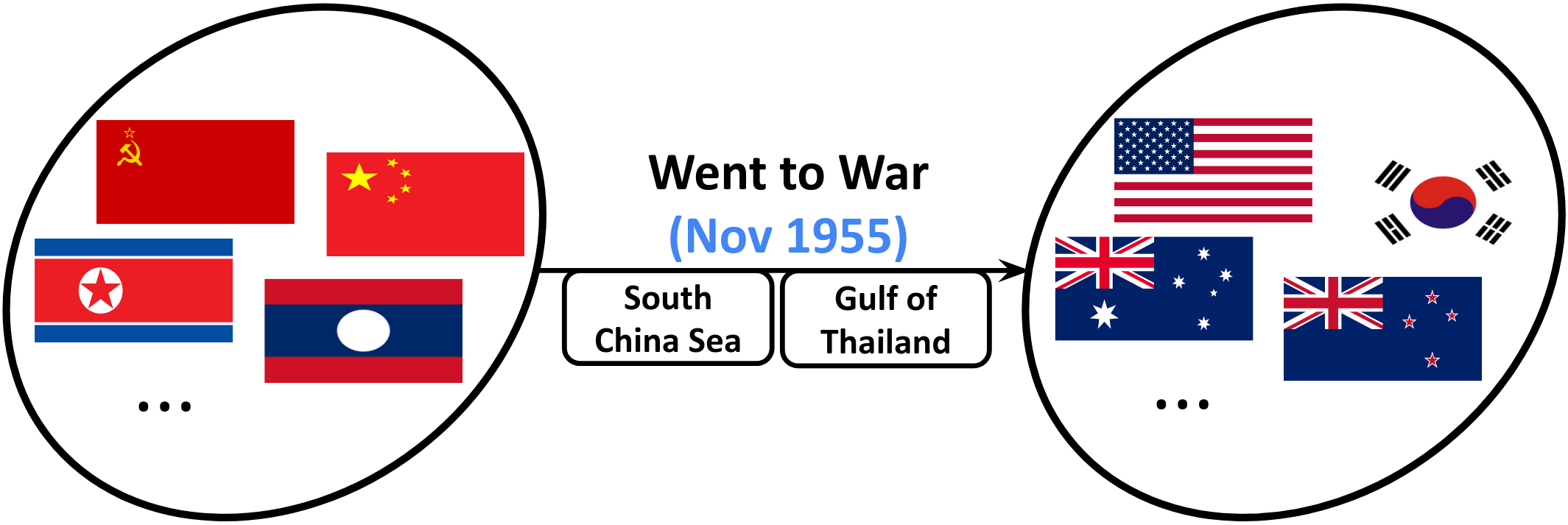}
    \caption{Vietnam War is an example of a complex geopolitical event that involved two coalitions of countries (\textit{i.e.,} two sets of entities) engaging in a war (\textit{i.e.,} an action) in locations such as the South China Sea and Gulf of Thailand (\textit{i.e.,} a set of qualifiers).}
\label{fig:actor_rec}
\end{figure}

\subsection{HTKG Limitations}
\label{sec:htkg_lim}
While HTKG extends TKG by including attributes/qualifiers, it still lacks natural support for efficiently expressing complex temporal facts involving more than two primary entities.
In this work, we highlight two types of such facts:

\paragraph{Group-Type} Facts involving more than two primary entities related to the same relation, which conceptually form a graph clique.
Typical examples of such facts include recurring political or financial summits that involve multiple nations attending simultaneously.
Expressing such an event in HTKG requires decomposing a singular fact into multiple facts and, in some cases, introducing new entities (\textit{i.e.,} reification).
However, these transformations are inefficient in terms of redundancy, sparsity, and even inference, as the predictor must aggregate information across many facts rather than a single one.
\autoref{fig:orig_vs_decomp} illustrates an example of an original fact concerning multiple entities along with the decomposed version adapted for HTKGs\footnote{\href{https://en.wikipedia.org/wiki/China-Japan-South_Korea_Free_Trade_Agreement}{China–Japan–South Korea Free Trade Agreement}}.
Moreover, if a new entity is not introduced to represent the event or the group of entities, which itself greatly amplifies the sparsity issues, HTKG struggles to distinguish between decomposed facts and separate co-occurring facts.

\paragraph{Set2Set-Type} Facts that involve a group of entities (\textit{i.e.,} actors) doing something to another group of entities (\textit{i.e.,} recipients), which conceptually form a biclique.
For example, in the geopolitical domain, numerous ongoing, emerging, and shifting coalitions form among different entities due to shared interests or other contemporary factors.
These coalitions engage with each other in various settings, such as economic or ideological warfare, causing complex events that require proper temporal structures for accurate representation.
Similar to the group-type, expressing such facts in HTKGs requires decomposing the original fact into multiple facts and/or introducing new entities, which leads to the same redundancy, sparsity, etc. issues.
\autoref{fig:actor_rec} illustrates an example of this type of fact concerning two groups of entities\footnote{\href{https://en.wikipedia.org/wiki/Vietnam_War}{Vietnam War}}.
Note that in this type of fact, at least one of the two groups of entities must be unnamed (\textit{i.e.,} unlike NATO and the European Union); otherwise, we can replace both groups with named entities representing them, which is a supported type of fact in HTKG.

While it is possible to extend the temporal structure further to support more complex and higher-order facts, such as those involving multiple groups of entities within one relation, given their rare occurrence in POLECAT and generally in real-world scenarios, we leave them for future investigations.

%% file: sections/4.htkgh.tex
\section{Hyper-Relational Temporal Knowledge Generalized Hypergraph (HTKGH)}
\label{sec:htkgh}

\paragraph{Hypergraph}
In contrast to regular graphs, edges can connect more than two vertices in hypergraphs, making them ideal to represent group-type facts.
By simply replacing regular graphs in HTKGs with hypergraphs, we arrive at the hyper-relational temporal knowledge hypergraph (HTKH) structure.
Formally, we define an HTKH $\mathcal{K}$ as
\begin{align}
     \resizebox{\columnwidth}{!}{$\mathcal{K} = \{ (\Gamma, Q) \; | \; \Gamma \in \mathbb{P}^+(\mathcal{E}) \times \mathcal{R} \times \mathcal{T}, Q \subseteq \mathcal{R} \times \mathcal{E} \}$} \nonumber
\end{align}
where $\mathbb{P}^+(\mathcal{E})$ is the non-null power set of entities.
While HTKHs can properly represent group-type facts, they still struggle with expressing the set2set-type facts.
Moreover, all edges in HTKHs are bidirectional, which is incompatible with existing unidirectional edges supported in HTKG.

\paragraph{Generalized Hypergraphs}
To address the shortcomings of HTKHs, we utilize a generalization of hypergraphs that allows second-order edges between first-order edges (\textit{i.e.,} edges in HTKH), deriving the hyper-relational temporal knowledge generalized hypergraph (HTKGH) structure.
By doing so, we reintroduce unidirectional edges to the structure and add support for set2set-type facts.
Formally, we define an HTKGH $\mathcal{W}$ as
\begin{align}
     \mathcal{W} = \{ (\Lambda, Q) \; | \; &\Lambda \in \mathbb{P}^+(\mathcal{E}) \times \mathcal{R} \times \mathbb{P}(\mathcal{E}) \times \mathcal{T},\nonumber\\
     &|\Lambda_\text{actors}| + |\Lambda_\text{recipients}| > 1,\nonumber\\
     &Q \subseteq \mathcal{R} \times \mathcal{E} \} \nonumber
\end{align}
where $\mathbb{P}(\mathcal{E})$ is the power set of entities.
In this formalization, we assume a special type of null first-order edge, $e_\varnothing$, that allows us to connect entities without any pre-determined relationship.
Moreover, we denote the first non-empty group of entities as ``\textit{actors}'' and the second group of entities as ``\textit{recipients}''.
Following these denotations, a group-type fact is expressed with an empty group of recipients, showcasing a relationship between the actors.
Moreover, a set2set-type fact is expressed with a non-empty group of recipients.

The derived HTKGH formalization can be used by both graph neural networks (GNNs) and LLMs approaches.
While our focus is mostly on studying LLMs, specifically how they perform on these highly complex structures, \autoref{sec:exp} and \autoref{app:gnn} provide a series of experiments and mechanisms to adapt GNNs.

%% file: sections/5.dataset.tex
\section{Dataset}

\label{sec:data}
POLitical Event Classification, Attributes, and Types (POLECAT)~\citep{halterman2023plover} is a new global events database built to succeed the Integrated Conflict Early Warning System (ICEWS)~\citep{DVN/28075_2015}.
POLECAT utilizes the Political Language Ontology for Verifiable Event Records (PLOVER) codebook for coding event data, a new event-mode-context ontology designed to be more general, easy to implement, and extendable, replacing the Conflict and Mediation Event Observations (CAMEO) framework~\citep{gerner2002conflict} utilized in ICEWS.
Compared to CAMEO, PLOVER defines 18 event types, aggregating over many of the more than 250 CAMEO codes.
To create \texttt{htkgh-polecat}, we utilize the POLECAT data from Jan 2018 to Jul 2024.
\autoref{app:data} provides more details on dataset construction and statistics, such as entity and relation composition.

\paragraph{Data Filtering}
To enforce HTKGH's requirements and reduce the noise in the samples, we remove all facts that have zero actors or have one actor and no recipients.
Doing so reduces the number of facts from 2.23M in the original database to approximately 556K.
While this is a significant reduction, such filtering ensures dataset integrity and enhances information quality.

\paragraph{Anonymous Variation}
One of the challenges of evaluating LLMs is dealing with information leaks.
Specifically, when the collected data is gathered from publicly accessible sources before the training cutoff date of an LLM, there is a chance that the model has seen that data during the training phase.
Given the highly public nature of political events, this problem is exacerbated in this domain.
As such, we create anonymous versions of \texttt{htkgh-polecat} by shuffling among entities and relations.
These variations enable us to evaluate LLMs' pattern recognition capabilities accurately 1) without compromising information leaks or memorization from pre-training and 2) in counterintuitive scenarios (see \autoref{app:anon} for more details).

\begin{table*}[t]
  \centering
  \resizebox{0.98\textwidth}{!}{
  \begin{tabular}{c|ccc|ccc|ccccc|cccc}
    \toprule
    \multirow{2}{*}{Variation} & \multicolumn{3}{c|}{Filters} & \multicolumn{3}{c|}{Heuristics} & \multicolumn{5}{c|}{Non-thinking} & \multicolumn{4}{c}{Thinking} \\
    \cmidrule{2-16}
    & Entity & Location & Context & $\bm{\mathcal{F}}$ & $\bm{\mathcal{R}}$ & $\bm{\mathcal{C}}$ & \textcolor{blue}{\textbf{L}} & \textcolor{violet}{\textbf{Q4N}} & \textcolor{violet}{\textbf{Q8N}} & \textcolor{red}{\textbf{G4}} & \textcolor{red}{\textbf{G12}} & \textcolor{SkyBlue}{\textbf{D}} & \textcolor{violet}{\textbf{Q4T}} & \textcolor{violet}{\textbf{Q8T}} & \textbf{O} \\
    \midrule
    \multirow{8}{*}{Regular}
    & True  & True  & True  & 40.2 & 29.4 & 19.4 & 19.9 & 33.8 & 28.8 & 28.0 & 36.8 & 13.7 & 36.7 & 35.9 & 34.8 \\
    & True  & True  & False & 42.5 & 28.9 & 15.4 & 24.5 & 35.9 & 29.9 & 31.5 & 39.1 & 12.7 & 39.1 & 38.9 & 34.8 \\
    & True  & False & True  & 42.9 & 30.3 & 16.6 & 23.2 & 36.3 & 30.8 & 30.8 & 39.6 & 13.1 & 39.0 & 38.3 & 36.1 \\
    & True  & False & False & 43.3 & 27.3 & 12.6 & 24.3 & 36.3 & 30.9 & 32.7 & 40.0 & 12.2 & 40.6 & 38.9 & 34.7 \\
    & False & True  & True  & 36.4 & 22.8 & 10.6 & 19.7 & 29.5 & 26.4 & 27.5 & 34.2 & 9.7 & 34.7 & 34.8 & 30.3 \\
    & False & True  & False & 31.5 & 19.8 & 6.7  & 20.2 & 27.3 & 24.1 & 28.0 & \textbf{32.6} & 7.5 & \textbf{31.9} & 31.3 & 28.2 \\
    & False & False & True  & 34.8 & 19.6 & 6.4  & 19.1 & 27.8 & 25.9 & 28.0 & 33.3 & 8.3 & 34.3 & 33.2 & 28.6 \\
    & False & False & False & 26.5 & 15.0 & 2.5  & 16.0 & 23.4 & 20.9 & \textbf{26.7} & \textbf{29.3} & 6.8 & \textbf{29.9} & \textbf{29.3} & 24.6 \\
    \midrule
    \multirow{8}{*}{\makecell{Shuffled\\(Entities)}}
    & True  & True  & True  & - & - & - & \cellcolor{Cherry!8}-0.3 & \cellcolor{PineGreen!8}+0.1 & \cellcolor{PineGreen!8}+0.1 & \cellcolor{PineGreen!8}+0.1 & \cellcolor{PineGreen!8}+0.6 & \cellcolor{PineGreen!8}+0.8 & \cellcolor{PineGreen!8}+0.2 & \cellcolor{PineGreen!8}+0.1 & \cellcolor{PineGreen!8}+0.3 \\
    & True  & True  & False & - & - & - & \cellcolor{PineGreen!8}+0.1 & \cellcolor{PineGreen!8}+0.1 & \cellcolor{Cherry!8}-0.1 & \cellcolor{PineGreen!8}+0.1 & \cellcolor{PineGreen!8}+0.7 & \cellcolor{Cherry!8}-0.2 & \cellcolor{Cherry!8}-0.3 & \cellcolor{Cherry!16}-1.0 & \cellcolor{PineGreen!8}+0.3 \\
    & True  & False & True  & - & - & - & \cellcolor{PineGreen!8}+0.2 & \cellcolor{PineGreen!8}+0.5 & \cellcolor{PineGreen!8}+0.8 & \cellcolor{PineGreen!8}+0.5 & \cellcolor{PineGreen!8}+0.4 & \cellcolor{PineGreen!8}+0.8 & \cellcolor{PineGreen!16}+1.0 & \cellcolor{PineGreen!8}+0.1 & \cellcolor{PineGreen!8}+0.3 \\
    & True  & False & False & - & - & - & \cellcolor{PineGreen!16}+1.1 & 0.0 & \cellcolor{PineGreen!8}+0.4 & \cellcolor{PineGreen!8}+0.6 & \cellcolor{PineGreen!8}+0.4 & \cellcolor{PineGreen!8}+0.6 & \cellcolor{Cherry!8}-0.3 & \cellcolor{Cherry!8}-0.8 & \cellcolor{PineGreen!8}+0.6 \\
    & False & True  & True  & - & - & - & \cellcolor{PineGreen!8}+0.4 & \cellcolor{PineGreen!16}+1.4 & \cellcolor{Cherry!8}-0.5 & \cellcolor{PineGreen!8}+0.4 & \cellcolor{PineGreen!8}+0.9 & \cellcolor{PineGreen!8}+0.6 & \cellcolor{PineGreen!8}+0.5 & \cellcolor{Cherry!8}-0.9 & \cellcolor{Cherry!8}-0.8 \\
    & False & True  & False & - & - & - & \cellcolor{Cherry!8}-0.3 & \cellcolor{PineGreen!8}+0.2 & \cellcolor{Cherry!8}-0.8 & \cellcolor{Cherry!8}-0.1 & \cellcolor{PineGreen!8}+0.3 & \cellcolor{PineGreen!16}+1.1 & \cellcolor{PineGreen!8}+0.2 & \cellcolor{Cherry!8}-0.1 & \cellcolor{Cherry!24}-2.2 \\
    & False & False & True  & - & - & - & \cellcolor{Cherry!8}-0.4 & \cellcolor{PineGreen!8}+0.6 & \cellcolor{Cherry!8}-0.7 & \cellcolor{Cherry!16}-1.2 & \cellcolor{PineGreen!8}+0.7 & \cellcolor{PineGreen!8}+0.3 & \cellcolor{Cherry!8}-0.2 & \cellcolor{Cherry!8}-0.3 & \cellcolor{Cherry!16}-1.3 \\
    & False & False & False & - & - & - & \cellcolor{Cherry!8}-0.8 & \cellcolor{Cherry!8}-0.3 & \cellcolor{Cherry!16}-1.2 & \cellcolor{Cherry!16}-1.4 & \cellcolor{PineGreen!8}+0.5 & \cellcolor{Cherry!8}-0.6 & \cellcolor{Cherry!8}-0.9 & \cellcolor{Cherry!16}-1.3 & \cellcolor{Cherry!24}-2.3 \\
    \midrule
    \multirow{8}{*}{\makecell{Shuffled\\(All)}}
    & True  & True  & True  & - & - & - & \cellcolor{PineGreen!32}+3.0 & \cellcolor{PineGreen!32}+3.2 & \cellcolor{PineGreen!56}+6.4 & \cellcolor{Cherry!32}-3.6 & \cellcolor{PineGreen!8}+0.2 & \cellcolor{PineGreen!64}+7.6 & \cellcolor{Cherry!16}-1.8 & \cellcolor{PineGreen!24}+2.1 & 0.0 \\
    & True  & True  & False & - & - & - & \cellcolor{PineGreen!32}+3.5 & \cellcolor{PineGreen!24}+2.9 & \cellcolor{PineGreen!56}+6.5 & \cellcolor{Cherry!48}-5.7 & \cellcolor{PineGreen!16}+1.9 & \cellcolor{PineGreen!64}+7.3 & \cellcolor{Cherry!24}-2.0 & \cellcolor{PineGreen!16}+1.0 & \cellcolor{PineGreen!16}+1.9 \\
    & True  & False & True  & - & - & - & \cellcolor{PineGreen!32}+3.6 & \cellcolor{PineGreen!24}+2.8 & \cellcolor{PineGreen!64}+7.3 & \cellcolor{Cherry!32}-3.8 & \cellcolor{PineGreen!8}+0.8 & \cellcolor{PineGreen!80}+9.2 & \cellcolor{Cherry!16}-1.2 & \cellcolor{PineGreen!16}+1.8 & \cellcolor{PineGreen!16}+1.7 \\
    & True  & False & False & - & - & - & \cellcolor{PineGreen!48}+5.3 & \cellcolor{PineGreen!32}+3.6 & \cellcolor{PineGreen!56}+6.2 & \cellcolor{Cherry!48}-5.8 & \cellcolor{PineGreen!16}+1.7 & \cellcolor{PineGreen!64}+7.8 & \cellcolor{Cherry!16}-1.6 & \cellcolor{PineGreen!16}+1.9 & \cellcolor{PineGreen!24}+2.2 \\
    & False & True  & True  & - & - & - & \cellcolor{PineGreen!32}+3.7 & \cellcolor{PineGreen!32}+3.6 & \cellcolor{PineGreen!48}+5.8 & \cellcolor{Cherry!40}-4.4 & \cellcolor{PineGreen!8}+0.8 & \cellcolor{PineGreen!56}+6.3 & \cellcolor{Cherry!24}-2.0 & \cellcolor{PineGreen!8}+0.5 & \cellcolor{Cherry!8}-0.2 \\
    & False & True  & False & - & - & - & \cellcolor{PineGreen!48}+5.7 & \cellcolor{PineGreen!32}+3.5 & \cellcolor{PineGreen!64}+7.2 & \cellcolor{Cherry!56}-6.1 & \cellcolor{PineGreen!16}+1.3 & \cellcolor{PineGreen!48}+5.3 & \cellcolor{Cherry!24}-2.2 & \cellcolor{PineGreen!8}+0.3 & \cellcolor{PineGreen!8}+0.2 \\
    & False & False & True  & - & - & - & \cellcolor{PineGreen!40}+4.9 & \cellcolor{PineGreen!48}+5.1 & \cellcolor{PineGreen!56}+6.3 & \cellcolor{Cherry!40}-4.9 & \cellcolor{PineGreen!8}+0.8 & \cellcolor{PineGreen!56}+6.6 & \cellcolor{Cherry!24}-2.5 & \cellcolor{PineGreen!16}+1.0 & \cellcolor{PineGreen!8}+0.9 \\
    & False & False & False & - & - & - & \cellcolor{PineGreen!56}+6.9 & \cellcolor{PineGreen!40}+4.1 & \cellcolor{PineGreen!72}+8.5 & \cellcolor{Cherry!56}-6.8 & \cellcolor{PineGreen!8}+0.5 & \cellcolor{PineGreen!40}+4.9 & \cellcolor{Cherry!32}-3.2 & 0.0 & \cellcolor{Cherry!8}-0.2 \\
    \bottomrule
  \end{tabular}
  }
  \caption{Relation prediction accuracy (\%) on \texttt{htkgh-polecat} variations. For each test sample, we include the most recent 100 facts (after filtering) as contextual information. In regular variation, bold values beat all the respective heuristics. In both shuffled variations, the numbers are reported as performance differences compared to their regular counterparts. \textbf{Legend:} $\bm{\mathcal{F}}$ $\rightarrow$ \texttt{Frequency}, $\bm{\mathcal{R}}$ $\rightarrow$ \texttt{Recency}, $\bm{\mathcal{C}}$ $\rightarrow$ \texttt{Copy}, \textcolor{blue}{\textbf{L}} $\rightarrow$ \texttt{Llama-3.1-8B-Instruct}, \textcolor{violet}{\textbf{Q4N}} $\rightarrow$ \texttt{Qwen3-4B-Instruct-2507}, \textcolor{violet}{\textbf{Q8N}} $\rightarrow$ \texttt{Qwen3-8B (non-thinking)}, \textcolor{red}{\textbf{G4}} $\rightarrow$ \texttt{gemma-3-4b-it}, \textcolor{red}{\textbf{G12}} $\rightarrow$ \texttt{gemma-3-12b-it}, \textcolor{SkyBlue}{\textbf{D}} $\rightarrow$ \texttt{DeepSeek-R1-Distill-Qwen-7B}, \textcolor{violet}{\textbf{Q4T}} $\rightarrow$ \texttt{Qwen3-4B-Thinking-2507}, \textcolor{violet}{\textbf{Q8T}} $\rightarrow$ \texttt{Qwen3-8B (thinking)}, and \textbf{O} $\rightarrow$ \texttt{openai/gpt-oss-20b (medium reasoning effort)}.}
  \label{tab:main}
\end{table*}

%% file: sections/6.experiments.tex
\section{Experiments}
\label{sec:exp}

In these experiments, we mostly analyze the relation prediction task, which is similar to the common logical reasoning tasks for LLMs.

\subsection{Evaluation Setup}
\paragraph{Test Set}
To test our models, we first filter out samples in 2018 to ensure the existence of sufficient context and then create a stratified dataset along temporal and relational axes, sampling a 1\% test set (\textit{i.e.,} approximately 5.5k facts).

\paragraph{Historical Context}
To build a historical context of size $h$ for a given test fact $f$, we first use a combination of the following filters:
\begin{enumerate}
    \item \textbf{Entity Filter:} Filter out facts that do not have any primary entity in common with the primary entities of $f$.
    \item \textbf{Location Filter:} If the location is available in $f$, filter out facts that do not have the same location.
    \item \textbf{Context Filter:} If there is any context available in $f$, filter out facts that do not share at least one context with $f$.
\end{enumerate}
Then, we take the $h$ most recent facts that have a timestamp strictly smaller than $f$.
\autoref{app:prompt} provides the details of the prompt creation process.

\begin{table*}[t]
  \centering
  \resizebox{\textwidth}{!}{
  \begin{tabular}{c|ccc|ccc|cccc|ccccc|cccc}
    \toprule
    \multirow{2}{*}{Variation} & \multicolumn{3}{c|}{Filters} & \multicolumn{3}{c|}{Heuristics} & \multicolumn{4}{c|}{GNN} & \multicolumn{9}{c}{LLM} \\
\cmidrule{2-20}
    & Entity & Location & Context & $\bm{\mathcal{F}}$ & $\bm{\mathcal{R}}$ & $\bm{\mathcal{C}}$ & $\mathbf{B}_{A}$ & $\mathbf{B}_{M}$ & $\mathbf{Hy}_{A}$ & $\mathbf{Hy}_{M}$ & \textcolor{blue}{\textbf{L}} & \textcolor{violet}{\textbf{Q4N}} & \textcolor{violet}{\textbf{Q8N}} & \textcolor{red}{\textbf{G4}} & \textcolor{red}{\textbf{G12}} & \textcolor{SkyBlue}{\textbf{D}} & \textcolor{violet}{\textbf{Q4T}} & \textcolor{violet}{\textbf{Q8T}} & \textbf{O} \\
    \midrule
    \multirow{8}{*}{Regular} & True & True & True & 46.7 & 34.7 & 21.7 & \textbf{48.2} & \textbf{49.1} & \textbf{48.8} & \textbf{49.1} & 23.0 & 38.2 & 32.4 & 31.4 & 40.6 & 12.2 & 41.0 & 39.7 & 39.3 \\
     & True & True & False & 46.4 & 31.2 & 17.4 & \textbf{48.7} & \textbf{48.9} & \textbf{49.4} & \textbf{49.0} & 26.8 & 38.9 & 32.4 & 33.8 & 42.2 & 11.7 & 41.7 & 42.0 & 38.0 \\
     & True & False & True & 46.9 & 33.1 & 18.6 & \textbf{48.5} & \textbf{48.6} & \textbf{48.2} & \textbf{48.7} & 25.4 & 39.5 & 34.3 & 33.6 & 42.8 & 12.2 & 42.8 & 41.5 & 39.9 \\
     & True & False & False & 46.2 & 28.8 & 14.2 & \textbf{48.2} & \textbf{49.4} & \textbf{48.2} & \textbf{48.2} & 26.7 & 38.7 & 33.9 & 34.8 & 43.0 & 12.2 & 43.4 & 41.4 & 38.6 \\
     & False & True & True & 40.8 & 26.0 & 12.4 & \textbf{48.2} & \textbf{49.1} & \textbf{49.3} & \textbf{48.7} & 22.0 & 33.6 & 29.6 & 30.4 & 37.1 & 9.3 & 38.2 & 37.4 & 33.2 \\
     & False & True & False & 33.3 & 21.0 & 8.0 & \textbf{49.1} & \textbf{48.4} & \textbf{48.9} & \textbf{48.8} & 23.1 & 30.2 & 26.9 & 31.0 & \textbf{35.3} & 8.0 & \textbf{34.9} & \textbf{35.0} & 31.2 \\
     & False & False & True & 37.3 & 22.0 & 7.6 & \textbf{49.1} & \textbf{49.2} & \textbf{49.5} & \textbf{49.1} & 21.8 & 31.2 & 28.6 & 29.6 & 36.6 & 8.1 & 36.6 & 35.5 & 30.6 \\
     & False & False & False & 25.0 & 15.0 & 2.4 & \textbf{49.8} & \textbf{49.4} & \textbf{49.2} & \textbf{49.7} & 18.8 & \textbf{26.1} & 23.8 & \textbf{28.5} & \textbf{32.5} & 6.4 & \textbf{32.8} & \textbf{31.1} & \textbf{26.9} \\
    \midrule
  \end{tabular}
  }
  \caption{Relation prediction accuracy (\%) on the last 24 months of the \texttt{htkgh-polecat} test set. For each test sample, an LLM predictor is provided with the most recent 100 facts after filtering, whereas a GNN predictor is provided with the last 4 windows. Bold values beat all the respective heuristics. \textbf{Legend:} $\bm{\mathcal{F}}$ $\rightarrow$ \texttt{Frequency}, $\bm{\mathcal{R}}$ $\rightarrow$ \texttt{Recency}, $\bm{\mathcal{C}}$ $\rightarrow$ \texttt{Copy}, $\mathbf{B}_{A}$ $\rightarrow$ \texttt{Bagging aggregation with mean pooling}, $\mathbf{B}_{M}$ $\rightarrow$ \texttt{Bagging aggregation with max pooling}, $\mathbf{Hy}_{A}$ $\rightarrow$ \texttt{Hypergraph aggregation with mean pooling}, $\mathbf{Hy}_{M}$ $\rightarrow$ \texttt{Hypergraph aggregation with max pooling}, \textcolor{blue}{\textbf{L}} $\rightarrow$ \texttt{Llama-3.1-8B-Instruct}, \textcolor{violet}{\textbf{Q4N}} $\rightarrow$ \texttt{Qwen3-4B-Instruct-2507}, \textcolor{violet}{\textbf{Q8N}} $\rightarrow$ \texttt{Qwen3-8B (non-thinking)}, \textcolor{red}{\textbf{G4}} $\rightarrow$ \texttt{gemma-3-4b-it}, \textcolor{red}{\textbf{G12}} $\rightarrow$ \texttt{gemma-3-12b-it}, \textcolor{SkyBlue}{\textbf{D}} $\rightarrow$ \texttt{DeepSeek-R1-Distill-Qwen-7B}, \textcolor{violet}{\textbf{Q4T}} $\rightarrow$ \texttt{Qwen3-4B-Thinking-2507}, \textcolor{violet}{\textbf{Q8T}} $\rightarrow$ \texttt{Qwen3-8B (thinking)}, and \textbf{O} $\rightarrow$ \texttt{openai/gpt-oss-20b (medium reasoning effort)}.}
  \label{tab:main_gnn}
\end{table*}

\paragraph{Models}
We select 9 models from common publishers such as Google and Qwen, based on size and type, enabling further analysis on these critical factors (see \autoref{app:model} for more details).
For size, our models range from 4B to 20B.
For type, we focus on two common types: non-thinking (\textit{i.e.,} instruction-tuned) and thinking.
Finally, to better analyze their performance, we allow thinking models to generate up to 16384 tokens while non-thinking models are limited to 14 tokens (\textit{i.e.,} max length of the relations, nine, plus five), all using their preferred decoding strategy and parameters.

\subsection{Experimental Results}
\paragraph{Q1. How good are LLMs at relation prediction?}
To better understand LLMs' performance, we compare them against three baselines that use simple heuristics that have been shown to either impose strong bias in LLMs~\citep{lee-etal-2023-temporal} or have strong performance in TKG forecasting benchmarks~\cite{10.24963/ijcai.2024/444}.
More specifically, given the historical context $\mathcal{H}$, we calculate 1) \textit{Frequency} as the relation with the most occurrences, 2) \textit{Recency} as the relation that comes last, and 3) \textit{Copy} as the existence of a contextual sample with the same non-temporal elements as the test sample.
\autoref{tab:main} (Regular variation) presents our experimental results on nine different state-of-the-art non-thinking and thinking LLMs.
As is evident, LLMs are only able to beat the heuristics in very few scenarios, which showcases the difficulty of the task and potential for future improvements.
Looking at the non-thinking category, \texttt{gemma-3-12b-it} and \texttt{Qwen3-4B-Instruct-2507} consistently beat their peers in their family, showcasing critical improvements with size increases (\texttt{gemma-3-4b-it} vs. \texttt{gemma-3-12b-it}) and alignment improvements\footnote{Based on the release card, it seems that \texttt{Qwen3-*-2507} models have gone through an improved post-training phase (vs. \texttt{Qwen3-*} models) rather than a full pre-training phase.} (\texttt{Qwen3-8B (non-thinking)} vs. \texttt{Qwen3-4B-Instruct-2507}).
Moreover, we see that with entity filtering, which leads to more relevant facts being present in the history, all models tend to improve, showcasing the potential gains of improving history retrievers, similar to the findings of~\citet{xia-etal-2024-chain}.
In the thinking category, we notice similar observations on the filtering; however, we notice two surprising results: 1) \texttt{DeepSeek-R1-Distill-Qwen-7B} does not perform well at all, which could partially be explained by misformatting rates (see \autoref{app:parser}), and 2) \texttt{Qwen-3} models beat the significantly larger \texttt{gpt-oss} model by substantial margins in many cases, showcasing the power of smaller well-trained models.
Moreover, we note a massive jump of \textbf{\textcolor{PineGreen}{+2.7-8.4\%}} from non-thinking to thinking variation of the \texttt{Qwen3} models, which highlights the effectiveness of test-time scaling for the relation prediction task on TKGs.

\begin{figure}[t]
    \centering
    \includegraphics[width=\linewidth]{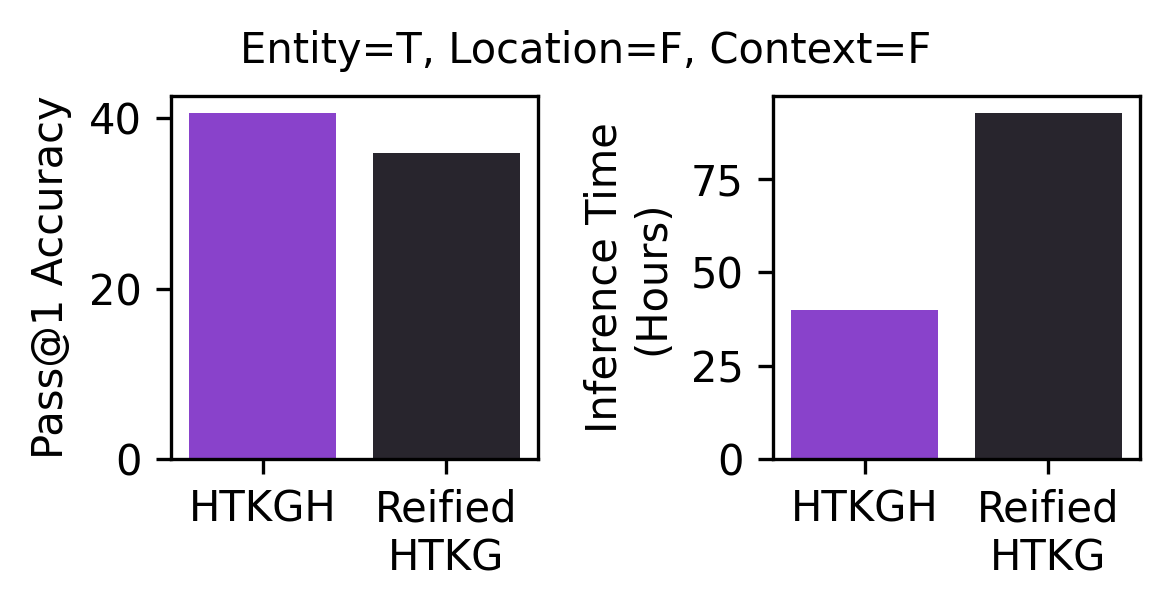}
    \caption{Relation prediction accuracy (\%) and inference time comparison on \texttt{Qwen3-4B-Thinking-2507} between HTKGH and reified HTKG.}
    \label{fig:reif}
\end{figure}

\paragraph{Q2. How much do LLMs rely on memorization for their predictions?}
While the observed improvements using test-time scaling are a sign of LLMs' going beyond memorization, we further experiment with two shuffled variations of our dataset, where 1) only countries are shuffled, and 2) countries and relations are shuffled.
\autoref{tab:main} (Shuffled variations) presents our experimental results for this experiment.
In the entity-only variation, we see that generally non-thinking models' accuracy slightly increases (up to \textbf{\textcolor{PineGreen}{+1.4\%}}) while thinking models' accuracy slightly decreases (up to \textbf{\textcolor{Cherry}{-2.3\%}}), which showcases their resilience and context-dependence to a great degree, similar to the findings of \citet{lee-etal-2023-temporal}.
However, when we also shuffle the relations, the behavior of many of the models becomes erratic, with some gaining up to \textbf{\textcolor{PineGreen}{+9.2\%}} accuracy and some losing up to \textbf{\textcolor{Cherry}{-6.8\%}} accuracy.
This phenomenon highlights the importance of testing LLMs under extreme settings where the provided information goes against the model's beliefs to different degrees.

\begin{figure*}[t]
    \centering
    \begin{subfigure}{0.49\linewidth}
        \centering
        \includegraphics[width=\linewidth]{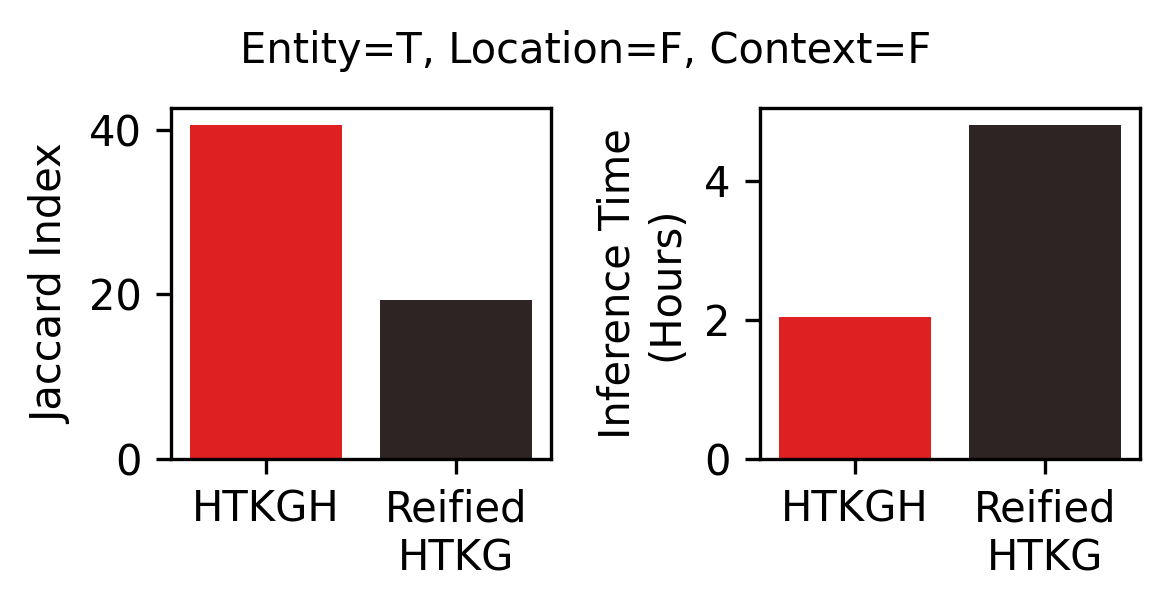}
        \caption{\texttt{gemma-3-12b-it}}
        \label{fig:gemma_size_effect}
    \end{subfigure}
    \begin{subfigure}{0.49\linewidth}
        \centering
        \includegraphics[width=\linewidth]{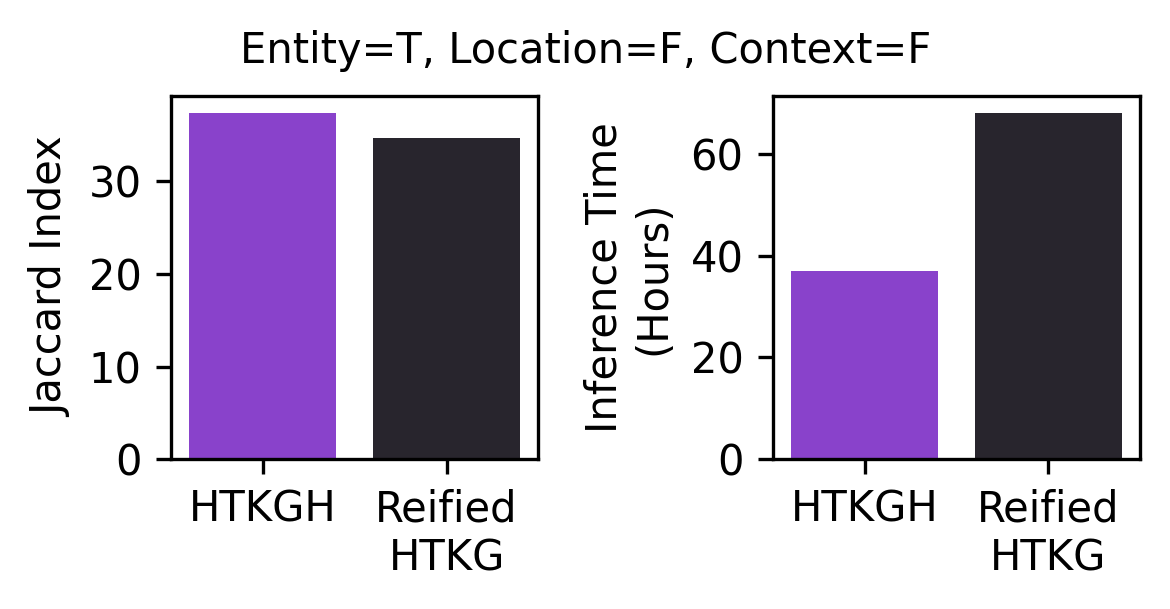}
        \caption{\texttt{Qwen3-4B-Thinking-2507}}
        \label{fig:qwen_size_effect}
    \end{subfigure}
    \caption{Link prediction performance (\textit{i.e.,} Jaccard Index) and inference time on \texttt{htkgh-polecat} between HTKGH and reified HTKG using the best non-thinking and thinking models, respectively.}
    \label{fig:reif_lp}
\end{figure*}

\paragraph{Q3. How good are LLMs compared to supervised graph-based models?}
We experiment with two GNN-based models, both based on first encoding fact-level information, then aggregating facts within a time window with either a permutation invariant \textit{bagging} or a more expressive \textit{hypergraph} approach, and finally using a transformer-based encoder to capture temporal trends (see \autoref{app:gnn} for more granular details on the GNN-based models).
Since these supervised models need a training phase, we split the dataset into a train (\textit{i.e.,} 2018-2022) and a test (\textit{i.e.,} 2023-2024) set.
Then, we randomly sample and hold out 10\% of the training data as a validation set.
Finally, we calculate the performance of the model on the validation set after each epoch and select the version with the best validation result for testing.
\autoref{tab:main_gnn} presents our experimental results comparing GNN-based models with our baselines and selected LLMs.
As is evident, in all scenarios, GNNs consistently beat LLMs, which we can potentially attribute to the information stored in embeddings and learned weights.
However, as we tighten our filtering, LLMs start to gain momentum, reducing the gap to supervised GNN-based models, whose performance stays flat, and getting as close as \textbf{\textcolor{Cherry}{-6\%}}.
These results emphasize the importance of retrieving high-quality context for predictions again.
Regarding our GNN-based models, we hypothesize that the current window encoding formulation introduces an information bottleneck by collapsing all fact representations, which leads to their flat performance across different filtered settings.
Given the numerous different modeling choices and designs for GNNs, we leave further investigations to future work.

\paragraph{Q4. How does HTKGH compare to HTKG?}
To better understand the direct impact of the new structure on the same underlying data, we compare HTKGH to reified HTKG using \texttt{htkgh-polecat}.
To this end, we reify each contextual sample by first adding a new node and then connecting each element to that node using a role-specific relation (\textit{e.g.,} actor, relation, etc.).
Moreover, for each query, we follow the same procedure but leave the relation edge empty so that the model can predict it.
This approach allows us to provide the model with the same amount of information as HTKGH samples during the test time, making the comparison fair.
\autoref{fig:reif} presents a comparison between HTKGH and reified HTKG on relation prediction accuracy and inference time.
As is evident, HTKGH beats reified HTKG by \textbf{\textcolor{PineGreen}{+4.7\%}} on performance and reduces inference time by \textbf{\textcolor{PineGreen}{$\sim$56.8\%}}. 
These results showcase the significant impact of utilizing HTKGH compared to more traditional common structures.
Note that we were not able to run decomposed HTKG~\citep{ding-etal-2024-temporal}, as our models ran out of context length (\textit{i.e.,} 32768 tokens) for 100 contextual samples.

\paragraph{Q5. How does HTKGH impact link prediction performance?}
While most of our experiments have been focused on relation prediction, to expand our insights into the impact of HTKGH, we compare it to reified HTKG on the link prediction task.
Given the structural shift from two-entity structures to set-based structures, we opt to use 1) a set-based prediction method (\textit{i.e.,} asking the model to predict a set of entities instead of one entity) and 2) the Jaccard Index instead of traditional single-entity recall-based metrics such as hits at $k$.
\autoref{fig:reif_lp} presents a comparison between HTKGH and reified HTKG on link prediction performance and inference time using \texttt{gemma-3-12b-it} and \texttt{Qwen3-4B-Thinking-2507}, the best non-thinking and thinking models, respectively.
As is evident, similar to our previous findings, beats reified HTKG by as much as \textbf{\textcolor{PineGreen}{21.29\%}} on performance and reduces inference time by \textbf{\textcolor{PineGreen}{$\sim$57.3\%}}.
These results once more highlight the critical benefits of utilizing HTKGH compared to existing common structures.

%% file: sections/7.conclusion.tex
\section{Conclusion}
In this work, we presented HTKGH, an extension of HTKG that is designed to efficiently and accurately express complex facts involving many primary entities.
Then, using the HTKGH's formalization, we introduced the \texttt{htkgh-polecast} dataset, built on top of the POLECAT event database.
Moreover, we conducted a thorough investigation into using state-of-the-art LLMs on relation prediction and link prediction tasks defined on \texttt{htkgh-polecast}.
Our results showcased 1) the out-of-the-box flexibility of LLMs to adapt to reasoning over complex higher-order facts and 2) the significant performance and inference time advantage HTKGH provides compared to existing structures, paving the way for future applications on even more complex structures and underlying data.

%% file: sections/99.appendix.tex
\section{Examples}
\label{app:exm}
\paragraph{Link Prediction Example}
Using a link prediction query, we can answer questions such as
``\unskip{
{\sethlcolor{VioletRed}\hl{\textit{Who}}} is going to 
{\sethlcolor{Goldenrod}\hl{\textit{win}}} the 
{\sethlcolor{SkyBlue}\hl{\textit{Super Bowl}}} in 
{\sethlcolor{SpringGreen}\hl{\textit{2026}}} at the 
{\sethlcolor{Thistle}\hl{\textit{Levi's stadium}}}?
}''
which is represented as
\begin{equation}
    ((\text{\sethlcolor{VioletRed}\hl{?}}, \text{\sethlcolor{Goldenrod}\hl{win}}, \text{\sethlcolor{SkyBlue}\hl{Super Bowl}}, \text{\sethlcolor{SpringGreen}\hl{2026}}),\{\text{\sethlcolor{Thistle}\hl{stadium: Levi's}}\}) \  .\nonumber
\end{equation}

\paragraph{Relation Prediction Example}
Using a relation prediction query, we can answer questions such as
``\unskip{
{\sethlcolor{Goldenrod}\hl{\textit{What}}} is going to happen between 
{\sethlcolor{VioletRed}\hl{\textit{Russia}}} and 
{\sethlcolor{SkyBlue}\hl{\textit{Ukraine}}} in 
{\sethlcolor{SpringGreen}\hl{\textit{2026}}} at 
{\sethlcolor{Thistle}\hl{\textit{Donbas}}}?
}''
which is represented as
\begin{equation}
    ((\text{\sethlcolor{VioletRed}\hl{Russia}}, \text{\sethlcolor{Goldenrod}\hl{?}}, \text{\sethlcolor{SkyBlue}\hl{Ukraine}}, \text{\sethlcolor{SpringGreen}\hl{2026}}),\{\text{\sethlcolor{Thistle}\hl{location: Donbas}}\}) \  .\nonumber
\end{equation}

\paragraph{Hyper-Relational Temporal Knowledge Hypergraph Example}
Using an HTKH, we can express facts such as
``\unskip{
{\sethlcolor{VioletRed}\hl{\textit{US}}}, 
{\sethlcolor{VioletRed}\hl{\textit{Canada}}}, and 
{\sethlcolor{VioletRed}\hl{\textit{Mexico}}} will 
{\sethlcolor{Goldenrod}\hl{\textit{negotiate}}} a new 
{\sethlcolor{Thistle}\hl{\textit{trade agreement}}} in 
{\sethlcolor{SpringGreen}\hl{\textit{2026}}}.
}''
in the form of
\begin{align}
    (&(\{\text{\sethlcolor{VioletRed}\hl{US}}, \text{\sethlcolor{VioletRed}\hl{Canada}}, \text{\sethlcolor{VioletRed}\hl{Mexico}}\}, \text{\sethlcolor{Goldenrod}\hl{negotiate}}, \text{\sethlcolor{SpringGreen}\hl{2026}}),\nonumber\\
    &\{\text{\sethlcolor{Thistle}\hl{type: trade agreement}}\}) \  .\nonumber
\end{align}

\paragraph{Hyper-Relational Temporal Knowledge Generalized Hypergraph Example}
Using an HTKGH, we can efficiently express facts such as
``\unskip{
{\sethlcolor{VioletRed}\hl{\textit{US}}} and 
{\sethlcolor{VioletRed}\hl{\textit{UK}}} 
{\sethlcolor{Goldenrod}\hl{\textit{sanction}}} 
{\sethlcolor{SkyBlue}\hl{\textit{Russia}}} and 
{\sethlcolor{SkyBlue}\hl{\textit{Belarus}}} in 
{\sethlcolor{SpringGreen}\hl{\textit{2022}}} over 
{\sethlcolor{Thistle}\hl{\textit{Russo-Ukrainian war}}}.
}''
in the form of
\begin{align}
    (&(\{\text{\sethlcolor{VioletRed}\hl{US}}, \text{\sethlcolor{VioletRed}\hl{UK}}\}, \text{\sethlcolor{Goldenrod}\hl{sanction}}, \{\text{\sethlcolor{SkyBlue}\hl{Russia}}, \text{\sethlcolor{SkyBlue}\hl{Belarus}}\}, \text{\sethlcolor{SpringGreen}\hl{2022}}),\nonumber\\
    &\{\text{\sethlcolor{Thistle}\hl{cause: Russo-Ukrainian war}}\}) \  .\nonumber
\end{align}

\section{Backward Compatibility}
It is trivial to show that any HTKG can be converted into an HTKGH, as we can replace each entity in the primary quadruple with a set containing only that entity.
Concretely, this entails the following conversion:
\begin{equation}
    ((s, r, o, t), Q) \rightarrow (\{s\}, r, \{o\}, t), Q) \nonumber
\end{equation}
Consequently, any model that can process HTKGHs will also be able to process HTKGs, which is crucial for maintaining backward compatibility with existing datasets.

\section{Dataset}
\label{app:data}
To create our dataset, rather than using raw text and parsing it ourselves, we utilize the already parsed data in the POLECAT database, which is stored in tabular format, and conform it to our HTKGH structure for our experiments.

\subsection{Dataset Construction}
\label{app:data_const}
\paragraph{Entity Construction}
To construct entities with more interactions (\textit{i.e.,} dense), we use countries instead of individuals (\textit{e.g.,} Vladimir Putin) as our entities.
Moreover, to add more specificity and resolution to the entities, whenever available, we include the sector information (\textit{e.g.,} judicial, government, or civilians).
This design choice creates multiple entities per country, each representing a slightly more granular actor with varying duties or interests.
For example, we do not include entities such as ``Mark Carney'' but instead use ``Canada (GOV)'' as the entity.
Ultimately, we end up with \textit{5268 entities}, built on top of 199 countries.

\paragraph{Relation Construction}
To construct the relations, whenever available, we combine the ``event type'' and `` event mode'' fields to provide more granularity regarding the actions in the facts.
One of the examples of such a merger is ``retreat (ceasefire)'' where the event mode (\textit{i.e.,} ceasefire) provides more resolution to the event type (\textit{i.e.,} retreat).
This results in \textit{42 relations} built from the original 18 event types in the PLOVER ontology.

\paragraph{Qualifier Construction}
We use two fields in the original coded events to construct the qualifiers: country and contexts.
When available, the former denotes the country where the event occurred.
The latter, when available, denotes the context of the event from 37 categories, such as military or legislative.
Since many events have multiple contexts, we add one qualifier for each.
In the end, each fact has, on average, \textit{1.37 qualifiers}.

\subsection{Dataset Statistics}
\label{app:data_stats}
\begin{figure*}[t]
    \centering
    \begin{subfigure}{0.32\linewidth}
        \includegraphics[width=\linewidth]{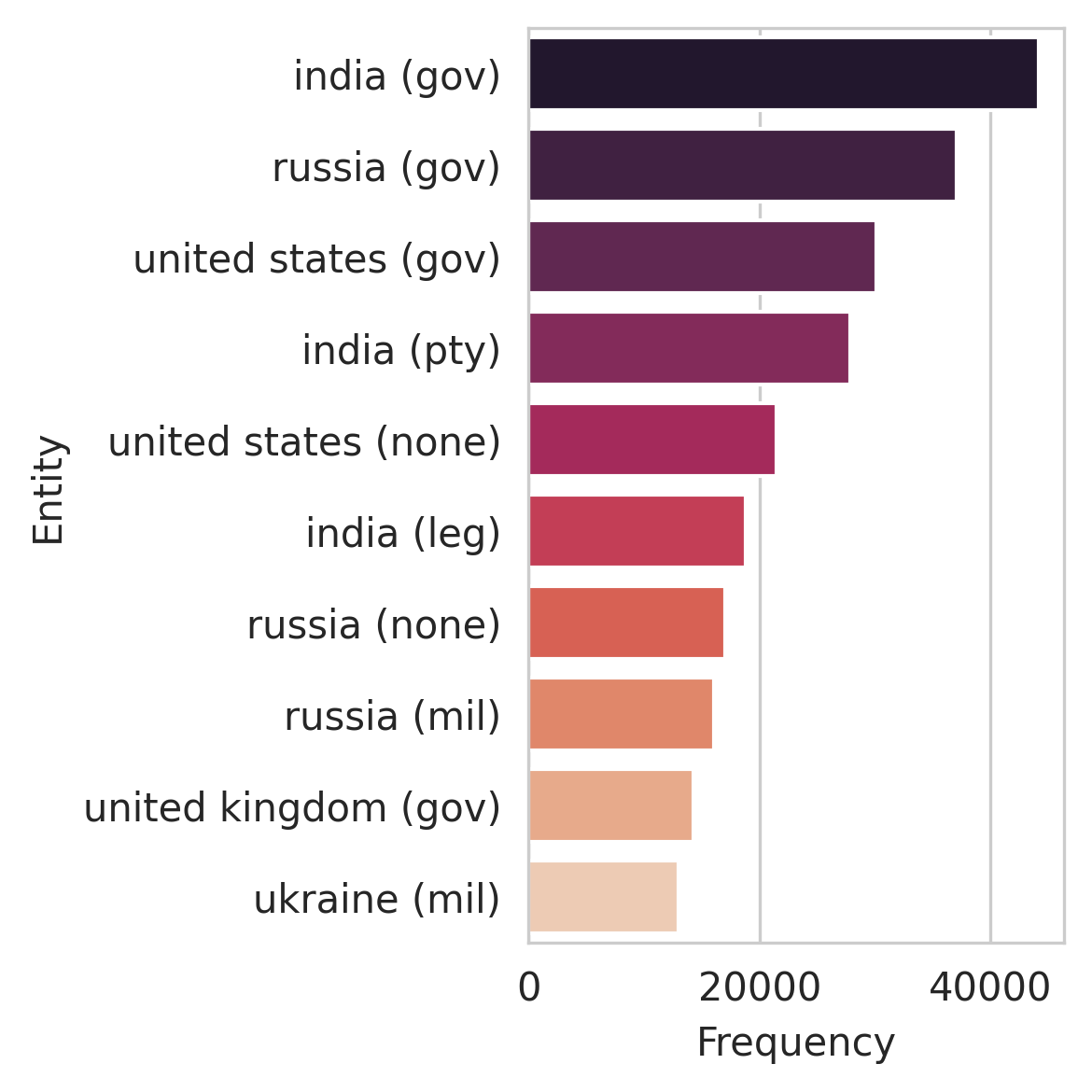}
        \caption{Top 10 Frequent Entities}
    \end{subfigure}
    \hspace*{\fill}
    \begin{subfigure}{0.32\linewidth}
        \includegraphics[width=\linewidth]{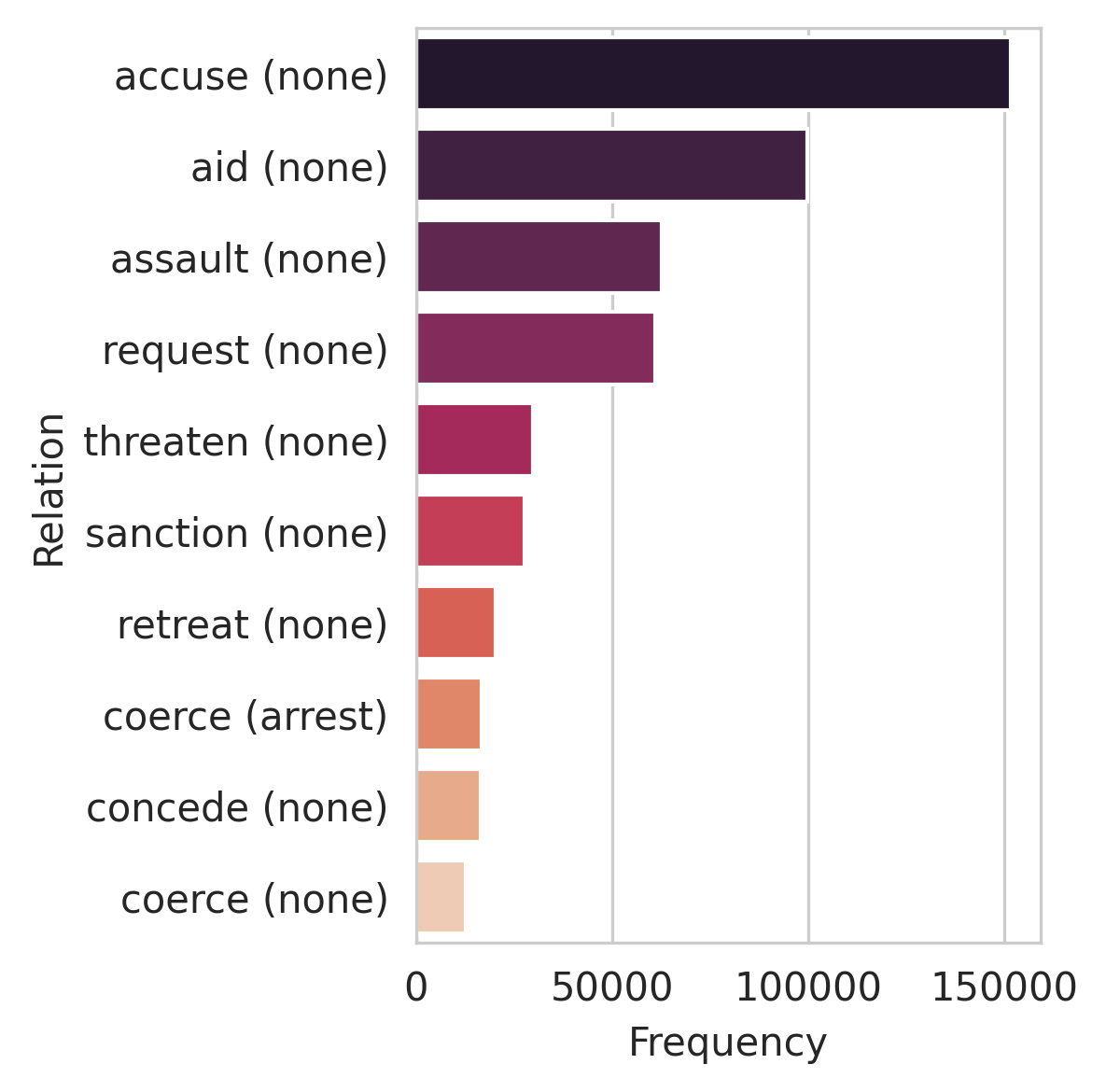}
        \caption{Top 10 Frequent Relations}
    \end{subfigure}
    \hspace*{\fill}
    \begin{subfigure}{0.32\linewidth}
        \includegraphics[width=\linewidth]{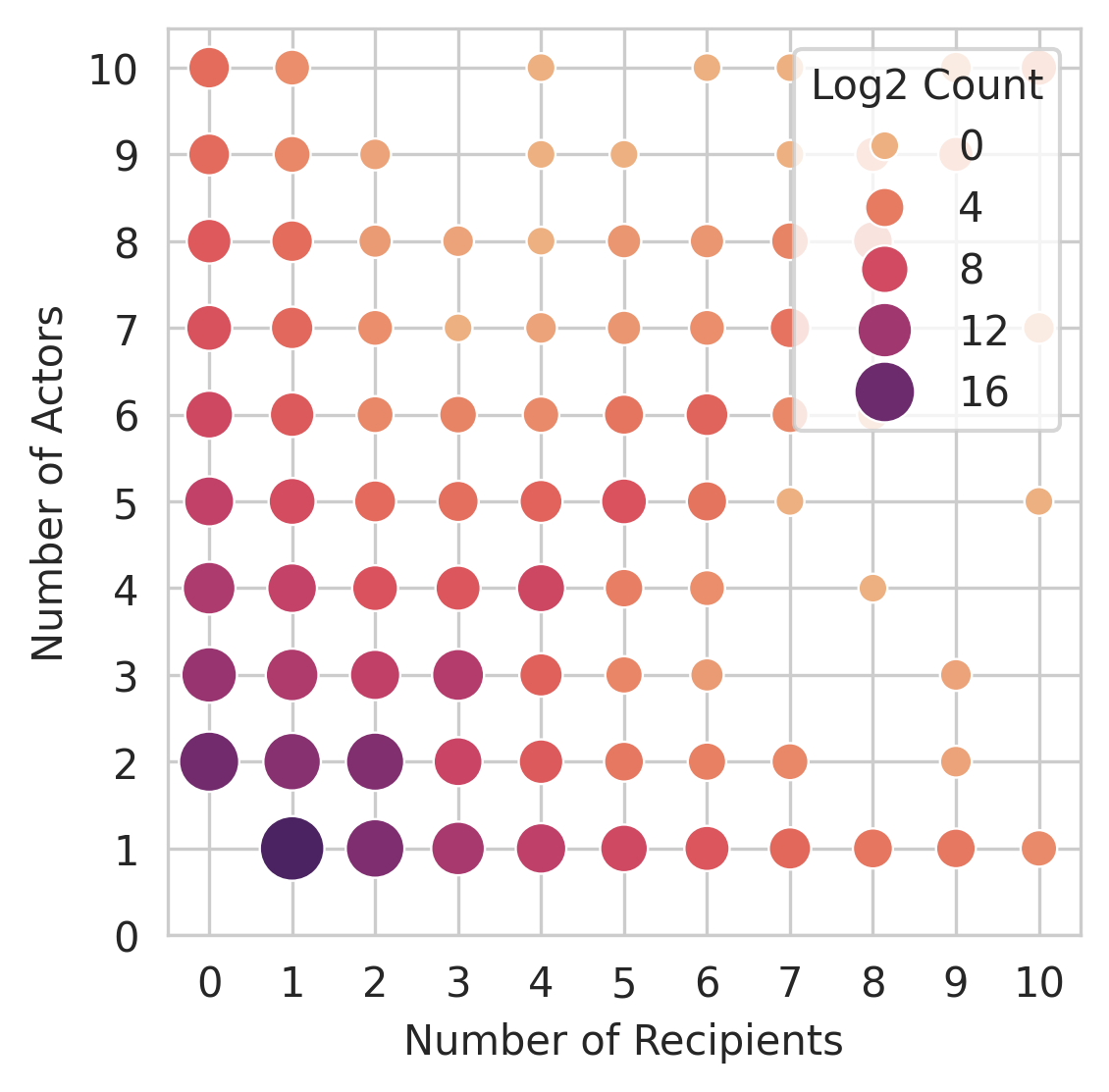}
        \caption{Number of Entities Frequency ($\le10$)}
    \end{subfigure}
    \caption{Statistics of the \texttt{htkgh-polecat} dataset: \textbf{(a)} The top 10 frequent entities. Each entity is comprised of a country and a sector within that country. \textbf{(b)} The top 10 frequent relations. Each relation is comprised of a type and a mode for that type. \textbf{(c)} The frequency of the number of entities (\textit{i.e.,} actors and recipients), up to 10 per group.}
    \label{fig:polecat_stats}
\end{figure*}

\begin{figure}[t]
    \centering
    \includegraphics[width=\linewidth]{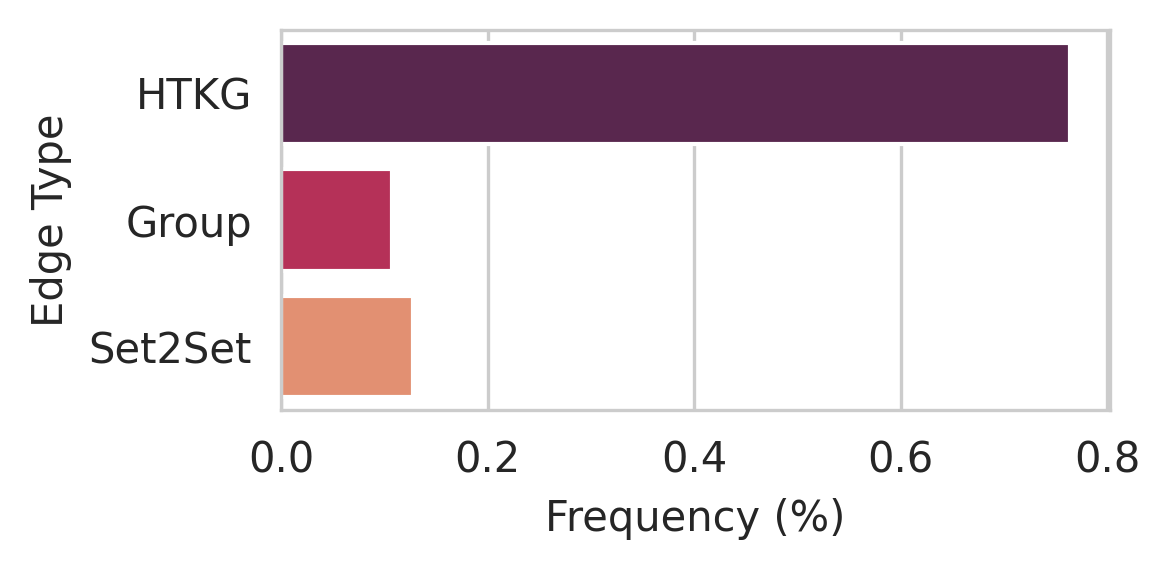}
    \caption{The frequency of different edge types in the \texttt{htkgh-polecat} dataset. The newly included edge types (\textit{i.e.,} Group and Set2Set) compose roughly 23.6\% of all edges, showcasing their prevalence in real-world data.}
\label{fig:polecat_et}
\end{figure}

\begin{figure}[t]
    \centering
    \includegraphics[width=\linewidth]{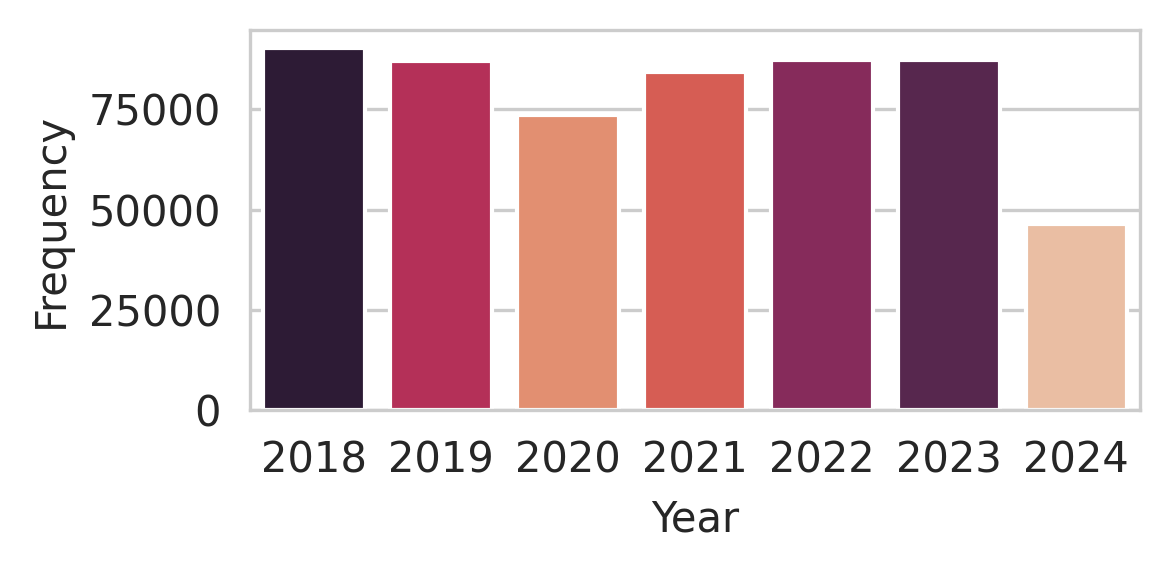}
    \caption{The frequency of events at different years in the \texttt{htkgh-polecat} dataset. The events are mostly distributed uniformly across years, except for 2024, where there is a roughly 46\% drop, due to a July 22 cutoff.}
\label{fig:polecat_time}
\end{figure}

\autoref{fig:polecat_stats} illustrates various statistics on the \texttt{htkgh-polecat} dataset, from the top frequent entities and relations to the distribution of the number of entities.
Moreover, \autoref{fig:polecat_et} presents the frequency of different edge types in the \texttt{htkgh-polecat} dataset.
As is evident, the two types of facts discussed in \autoref{sec:htkg_lim} encompass many facts in real-world political events.
Finally, \autoref{fig:polecat_time} showcases the distribution of events across different years, where we mostly see a uniform distribution across the years, except for 2024, where there is about a 46\% drop in the frequency of events due to a data cutoff in July.

\subsection{Test Set Construction}
\begin{figure*}[t]
    \centering
    \begin{subfigure}[t]{0.32\textwidth}
        \includegraphics[width=\linewidth]{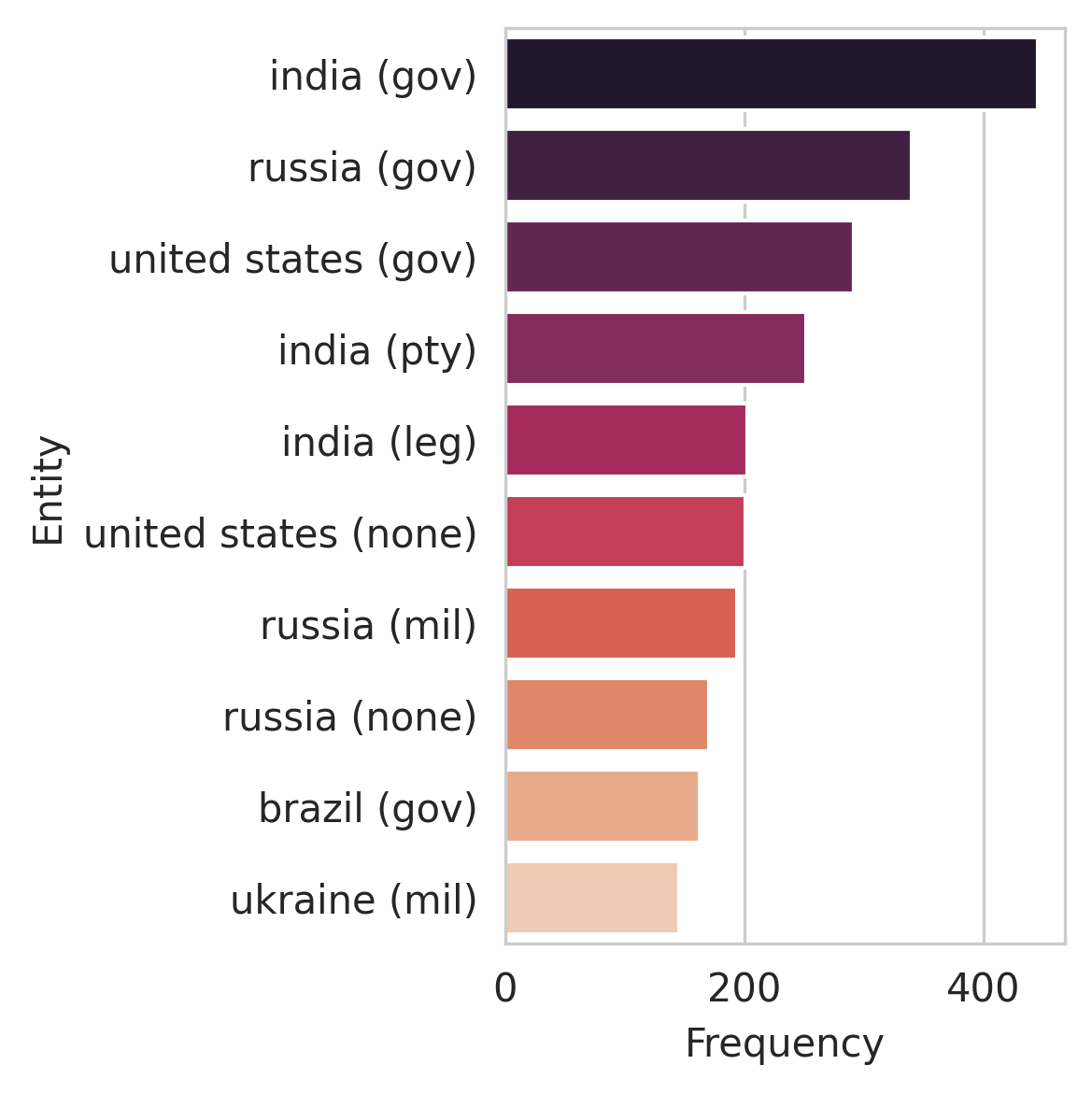}
        \caption{Top 10 Frequent Entities}
        \label{fig:polecat_stats_test_a}
    \end{subfigure}\hfill
    \begin{subfigure}[t]{0.32\textwidth}
        \includegraphics[width=\linewidth]{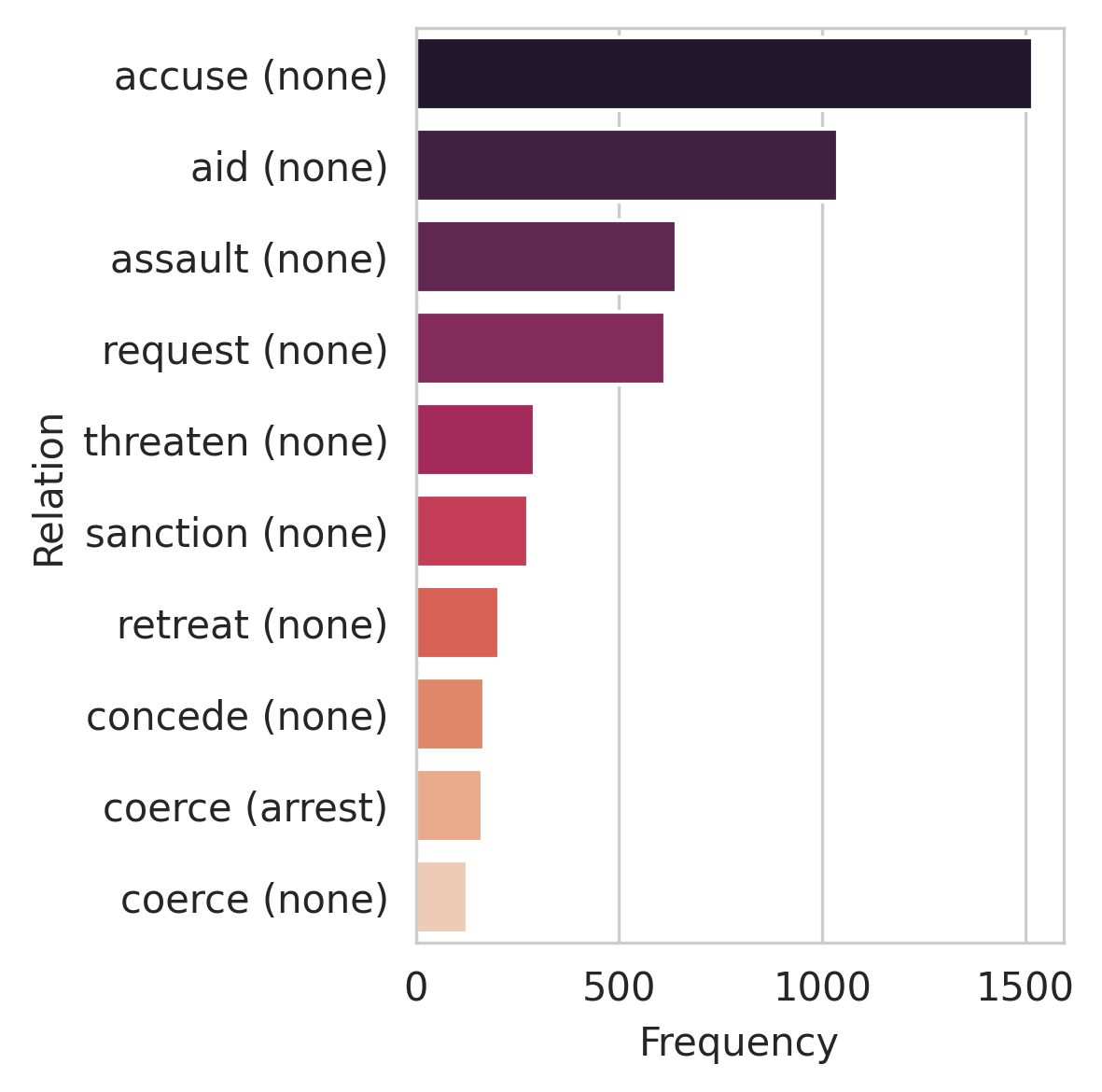}
        \caption{Top 10 Frequent Relations}
        \label{fig:polecat_stats_test_b}
    \end{subfigure}\hfill
    \begin{subfigure}[t]{0.32\textwidth}
        \includegraphics[width=\linewidth]{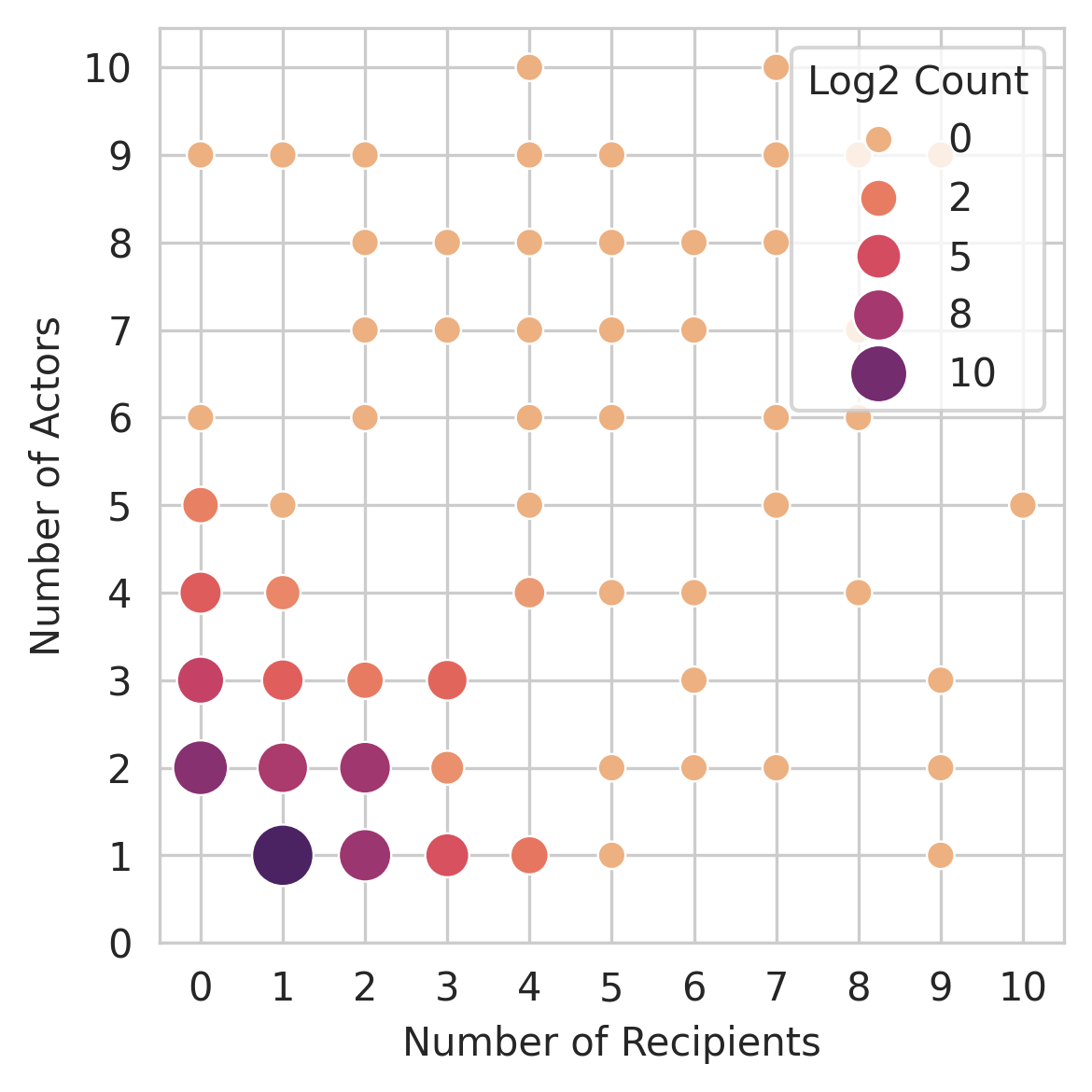}
        \caption{Number of Entities Frequency ($\le10$)}
        \label{fig:polecat_stats_test_c}
    \end{subfigure}

    \begin{subfigure}[t]{0.49\textwidth}
        \includegraphics[width=\linewidth]{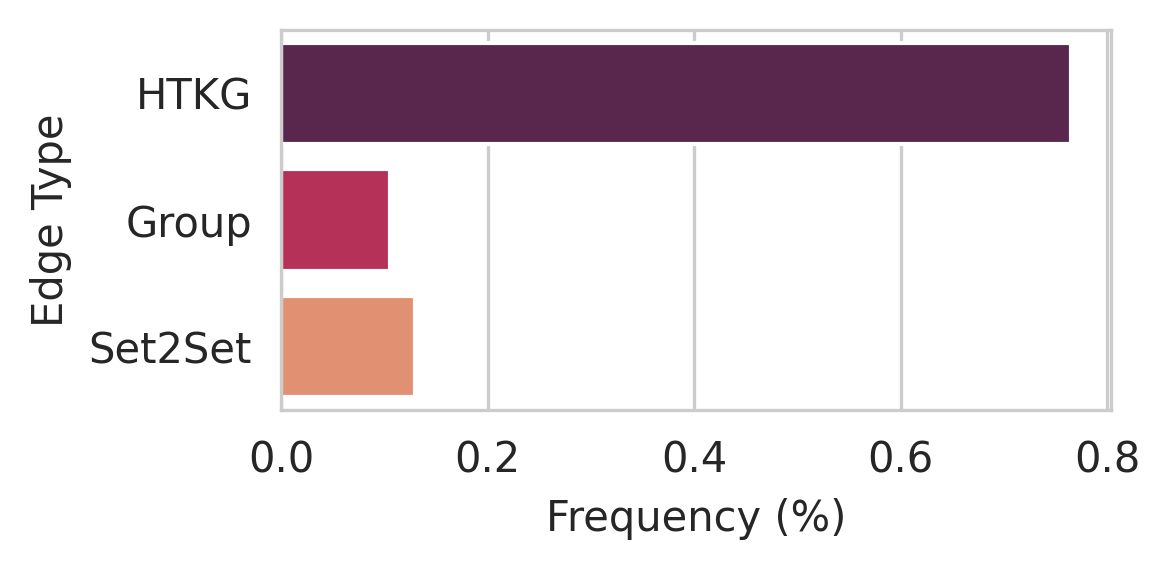}
        \caption{Edge Types}
        \label{fig:polecat_stats_test_d}
    \end{subfigure}\hfill
    \begin{subfigure}[t]{0.49\textwidth}
        \includegraphics[width=\linewidth]{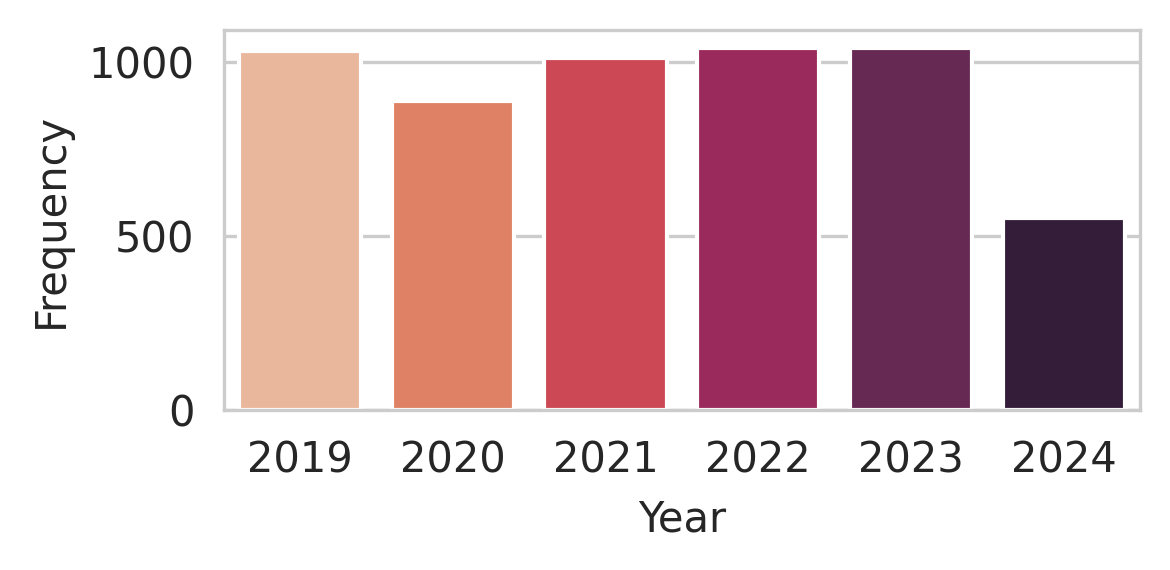}
        \caption{Sample by Year}
        \label{fig:polecat_stats_test_e}
    \end{subfigure}

    \caption{Statistics of the \texttt{htkgh-polecat} test set: 
    \textbf{(a)} The top 10 frequent entities. 
    \textbf{(b)} The top 10 frequent relations. 
    \textbf{(c)} The frequency of the number of entities (\textit{i.e.,} actors and recipients), up to 10 per group.
    \textbf{(d)} Edge type distribution.
    \textbf{(e)} Sample year distribution.
    Distributions are similar to those of the overall \texttt{htkgh-polecat} dataset, apart from having fewer facts with high numbers of actors or recipients, and the dropping of events from 2018 due to a lack of historical context in the dataset.}
    \label{fig:polecat_stats_test}
\end{figure*}

\autoref{fig:polecat_stats_test} presents descriptive statistics over the constructed test set.
As is evident, the most frequent entities are similar to those for the dataset overall (see \autoref{fig:polecat_stats}) with minor shuffling of the order outside the top-3.
Moreover, the distribution of relations, which were stratified upon, is unchanged from the overall dataset.
Similarly, the distribution over years is relatively unchanged, the only difference being that we drop 2018 from the test set to ensure the existence of a historical context for all the facts.
We do see fewer events with high numbers of actors or recipients, but they have not been eliminated from the test set.
Finally, as shown in \autoref{fig:polecat_et}, 23.5\% of the edges in the test set are of the newly included edge types, compared to 23.6\% in the overall dataset.

\section{Comparison to \texttt{tkgl-polecat}}
\label{app:comp}
Compared to \texttt{tkgl-polecat}, we 1) preserve all the primary entities, instead of choosing one on each side to force a TKG format, 2) include location and context qualifiers as extra information, 3) encompass a more extended period of time (2018-2023 vs. 2018-2024), 4) use a scheme to construct entities and relations, leading to a higher density in the graph, and 5) employ rigorous filtering to ensure quality of all the facts in the dataset.

\begin{figure*}[t]
    \centering
    \includegraphics[width=\linewidth]{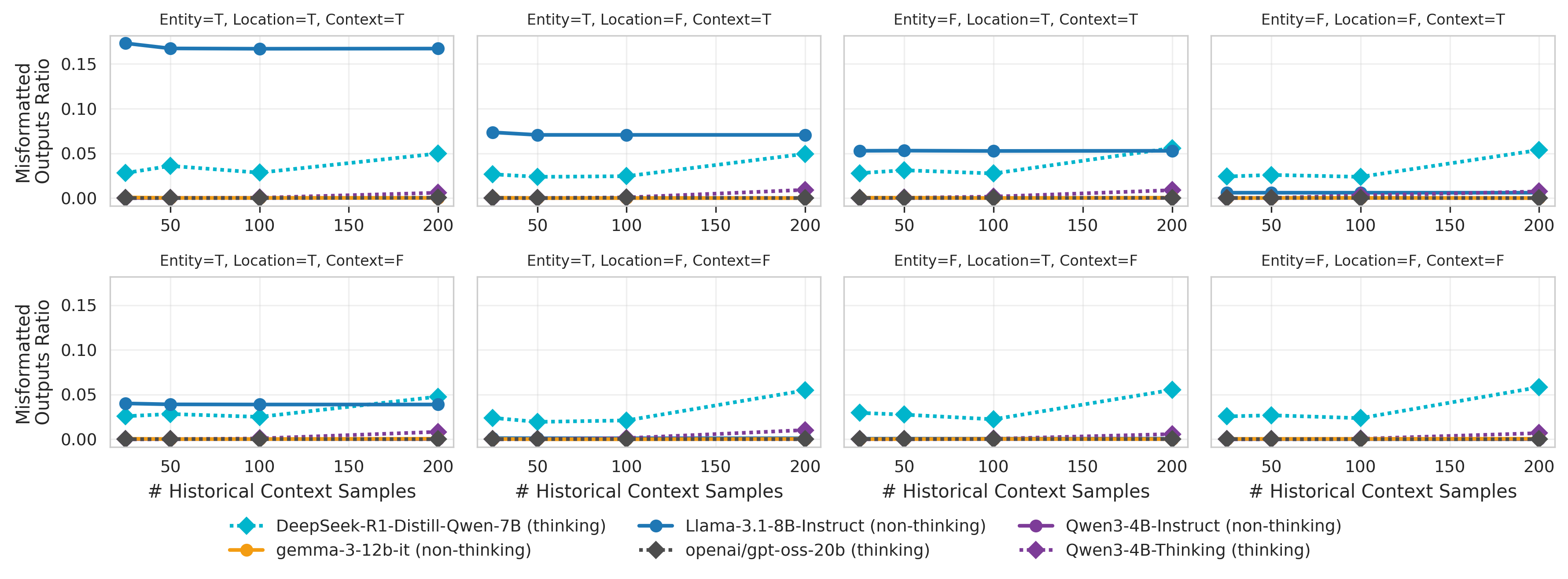}
    \caption{Proportion of misformatted outputs on \texttt{htkgh-polecat} variations. From each model family, we only display the best-performing member.}
    \label{fig:polecat_mf_frac}
\end{figure*}

\section{Construction of \texttt{htkgh-polecat-anon}}
\label{app:anon}
Accounting for information leaks from the pre-training phase, we create anonymized versions of \texttt{tkgl-polecat} to investigate memorization versus pattern recognition abilities properly.
To this end, we shuffle the entities and/or relations, practically transferring the facts to an imaginary world that the model has not seen before and is counterintuitive to its existing beliefs.
During anonymization, we ensure consistency by 1) prohibiting multiple original words from being shuffled with the same new word and 2) synchronizing the countries in the location qualifiers and the primary entities while shuffling.
Moreover, we consider the following three fields as potentially confounding information for anonymization: entities (actors and recipients), relations, and location qualifiers.
Finally, we create two variations based on the anonymized fields: 1) entities and country qualifiers only, 2) entities, country qualifiers, and relations.

\section{Prompt Templates}
\label{app:prompt}

\subsection{Non-thinking}
\label{app:prompt:non}

\begin{Verbatim}[frame=single, fontsize=\small]
You are given a series of related geopolitical
events, each described by its "actors",
"recipients", "relation", and "qualifiers".
Analyze the historical events and the new event
carefully, considering the context provided by
the historical events.
Based on your analysis, determine the most
likely "relation" from a list of candidates for
the new event.
For your final answer, only output the chosen
candidate in the same exact format as the
candidates without any additional text.

Output format example:
protest (strike)

Here are the historical events to analyze:
{context_samples}

Here is the new event to classify:
{fact}

Here are the candidate relations:
{candidates}

The most likely relation is:
\end{Verbatim}

\subsection{Thinking}
\label{app:prompt:thnk}

\begin{Verbatim}[frame=single, fontsize=\small]
You are given a series of related geopolitical
events, each described by its "actors",
"recipients", "relation", and "qualifiers".
Consider the new event and the provided context
carefully to determine the most likely
"relation" from a list of candidates for the
new event.
For your final answer, only output the chosen
candidate in the same exact format as the
candidates without any additional text.

Here are the historical events to analyze:
{context_samples}

Here is the new event to classify:
{fact}

Here are the candidate relations:
{candidates}
\end{Verbatim}

\begin{table}[t]
  \centering
  \resizebox{\columnwidth}{!}{
  \begin{tabular}{cccc}
    \toprule
    \textbf{Publisher} & \textbf{Model} & \textbf{\makecell{\#Params\\(B)}} & \textbf{Type} \\ \midrule

    \multirow{2}{*}{Google} 
    & \texttt{gemma-3-4b-it}    & 4.0   & N \\
    & \texttt{gemma-3-12b-it}   & 12.0  & N \\
    \midrule
    \multirow{4}{*}{Qwen} 
    & \texttt{Qwen3-4B-Instruct-2507}   & 4.0   & N \\
    & \texttt{Qwen3-4B-Thinking-2507}   & 4.0   & N \\
    & \texttt{Qwen3-8B}                 & 8.0   & H \\
    & \texttt{Qwen3-14B}                & 15.0  & H \\
    \midrule
    \multirow{1}{*}{Meta}
    & \texttt{Llama-3.1-8B-Instruct}    & 8.0   & N \\
    \midrule
    \multirow{1}{*}{DeepSeek}
    & \texttt{DeepSeek-R1-Distill-Qwen-7B}  & 8.0   & T \\
    \midrule
    \multirow{1}{*}{OpenAI}
    & \texttt{gpt-oss-20b}  & 22.0  & T \\
    \bottomrule
  \end{tabular}
  }
  \caption{We curate an inclusive list of thinking and non-thinking models from prominent model providers. \textbf{Legend:} \textbf{N} $\rightarrow$ Non-thinking, \textbf{T} $\rightarrow$ Thinking, and \textbf{H} $\rightarrow$ Hybrid.}
  \label{tab:models}
\end{table}

\begin{figure*}
    \centering
    \includegraphics[width=\linewidth]{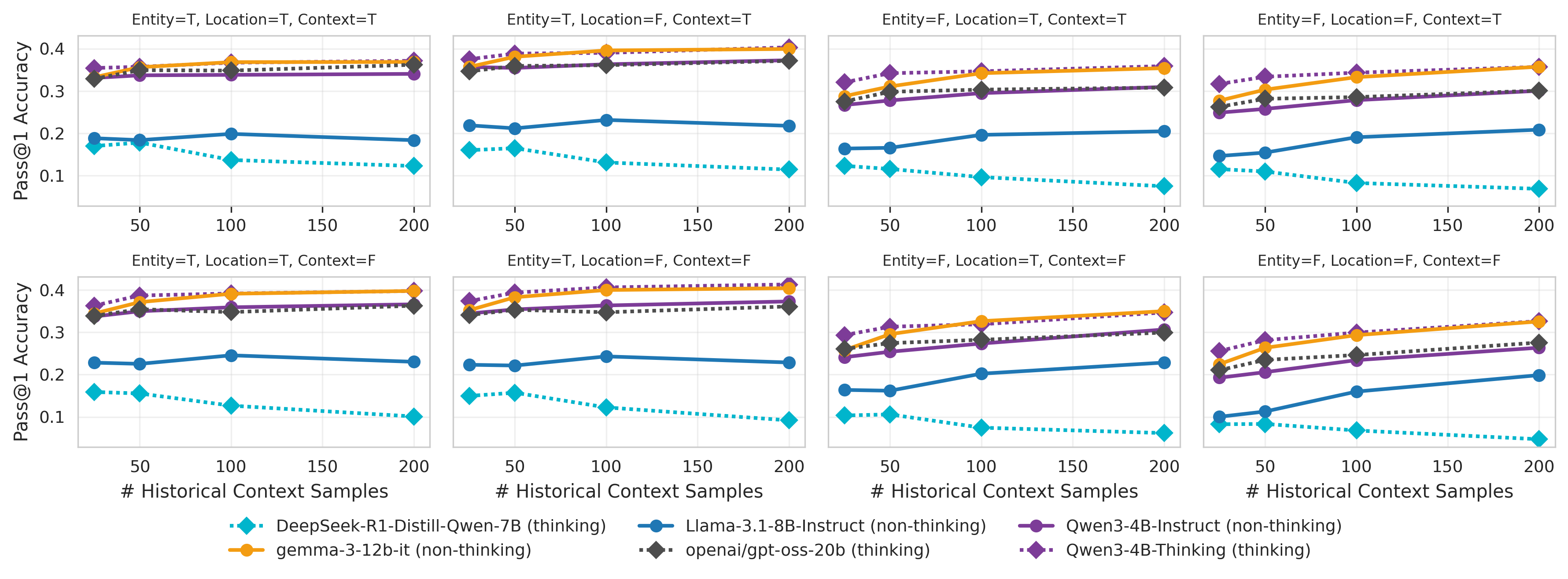}
    \caption{Relation prediction accuracy (\%) on \texttt{htkgh-polecat} context variations over the number of retrieved contextual samples. From each model family, we only display the best-performing member.}
    \label{fig:polecat_acc}
\end{figure*}

\begin{figure*}[t]
    \centering
    \includegraphics[width=\linewidth]{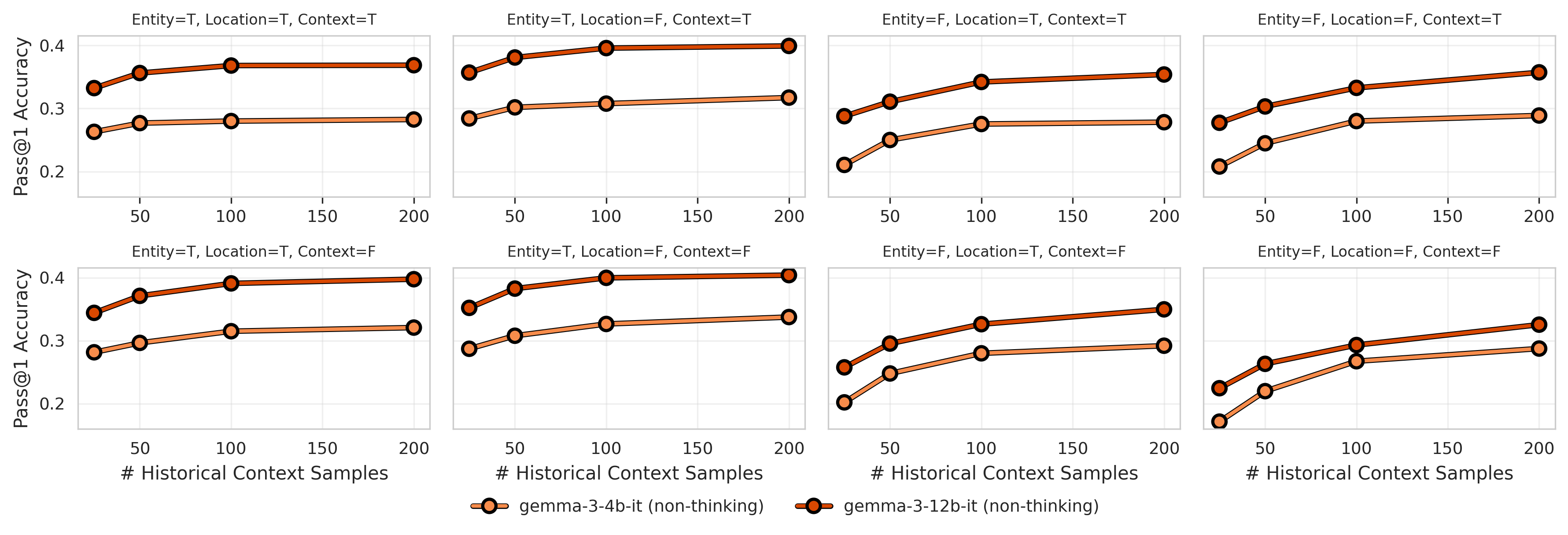}
    \caption{Relation prediction accuracy (\%) on \texttt{htkgh-polecat} variations for the Gemma-3 family.}
    \label{fig:app_model_size_gemma}
\end{figure*}

\begin{figure*}[t]
    \centering
    \includegraphics[width=\linewidth]{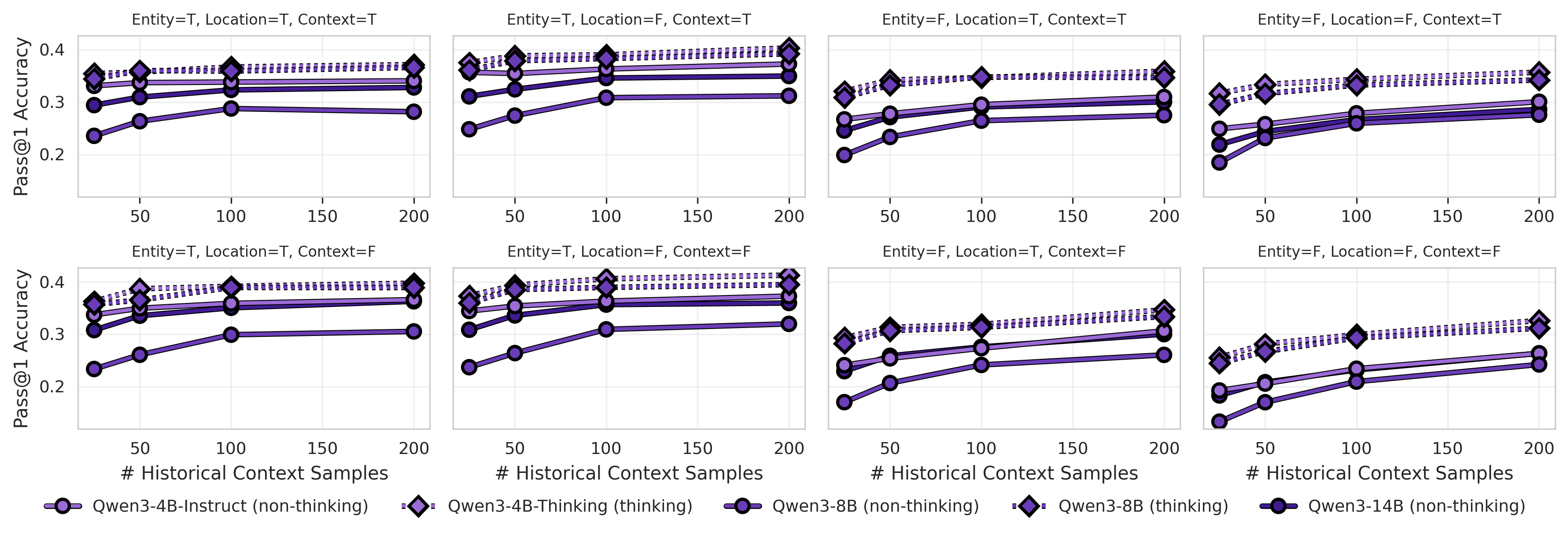}
    \caption{Relation prediction accuracy (\%) on \texttt{htkgh-polecat} variations for the Qwen3 family.}
    \label{fig:app_model_size_qwen}
\end{figure*}

\section{Models}
\label{app:model}
\autoref{tab:models} showcases all 9 models used in our experiments.

\section{Response Parser}
\label{app:parser}
To include as many responses as possible, we use a two-tier extractor that searches for matches on both the ground truth's raw text and index.
To this end, we extract all the possible answers that are in the correct format from the response, taking the first one as the final answer and comparing it to the ground truth's raw text.
If there are no matches, we extract all the numbers from the response, taking the first one as the final answer and comparing it to the ground truth's index.
\autoref{fig:polecat_mf_frac} shows the ratio of misformatted outputs that were not parsed by our method.
We can see that most models consistently exhibit a low rate of misformatted outputs, except for \texttt{Llama-3.1-8B-Instruct} and \texttt{DeepSeek-R1-Distill-Qwen-7B}.
Specifically, \texttt{DeepSeek-R1-Distill-Qwen-7B} increasingly struggles as we increase the number of contextual samples, while \texttt{Llama-3.1-8B-Instruct} specifically struggles when all filters are turned on.

\section{Implementation Details}
We implemented our codebase using the vLLM~\citep{kwon2023efficient} and Transformers~\citep{wolf-etal-2020-transformers} libraries.
All our experiments are run on a server with 8$\times$A6000 48GB GPUs and two servers with 8$\times$A5000 24GB GPUs.

\section{Unidirectional vs. Bidirectional Edges}
While traditionally HTKG facts are unidirectional, allowing group-type facts has a side effect of introducing bidirectional edges.
Specifically, since there is no inherent head or tail among the group of entities in these facts, the relation is semantically bidirectional, which enables us to express more diverse relations compactly within the same HTKGH structure.
Formally, the unidirectional ($r_u$) vs. bidrectional ($r_b$) relations manifest as follows:
\begin{equation}
    (\{v\}, r_u, \{u\}, t), Q) \;\; \text{vs.} \;\; (\{v, u\}, r_b, \{\}, t), Q) \nonumber
\end{equation}
Note that these bidirectional relations only apply to first-order edges, as second-order edges remain unidirectional in our formalization.

\section{Effect of Model Size}
\label{app:model_size}


While many training factors are often undisclosed, model size has always been one of the consistent predictors of performance.
\autoref{fig:app_model_size_gemma} and \autoref{fig:app_model_size_qwen} showcases a comparison with regard to model size within the Gemma-3 and Qwen3 families.
In both families, the performance generally increases as model sizes increase.
The most significant exceptions are the \texttt{Qwen3-4B-*-2507} models, which even beat the much larger \texttt{Qwen3-14B} model, underscoring the critical nature of other factors that must be studied for model selection.

\section{Effect of Number of Contextual Samples}
One of the critical considerations of using LLMs for TKG-related tasks is the number of facts that could be included in the context.
\autoref{fig:polecat_acc} illustrates our experimental results with respect to the number of contextual samples.
As is evident, most models generally benefit from the inclusion of more context, \texttt{DeepSeek-R1-Distill-Qwen-7B} being the only model that degrades with more samples.
Moreover, we observe that scenarios with stricter filtering benefit more from an increase in samples.
These results illustrate a promising trend in context scalings for more complex TKG-related predictions.
Note that some models allow us to scale the samples further, but due to resource constraints, we leave these experiments to future work.


\section{Graph Neural Network Models}
\label{app:gnn}

\subsection{Preliminaries}
Let a fact be defined as
\begin{equation}
    \psi = (A_{\psi}, r_{\psi}, R_{\psi}, t_{\psi}, Q_{\psi}),
\end{equation}
consisting of the actors set, relation label, recipients set, timestamp, and qualifier key-value pairs, respectively.
For a query fact $q$ at time $t_q$, we form a query-specific window history sequence with fixed lookback duration $\Delta$ and history length $\mathcal{H}$ as
\begin{equation}
    \mathcal{S}(q) = \{\mathcal{W}_0(q), \mathcal{W}_1(q), ..., \mathcal{W}_\mathcal{H}(q)\} \  .
\end{equation}
For each window index $j \in \{1, ..., \mathcal{H}\}$, facts in that window are defined as
\begin{equation}
    \begin{split}
    & \Psi _j(q) = \{ \\
    & ~~~~~ \psi: t_q-j\Delta \leq t_{\psi} < t_q-(j-1)\Delta ~ \land \\
    & ~~~~~ \kappa(\psi,q) = 1 \\
    & \},
    \end{split}
\end{equation}
where $\mathcal{W}_j(q)$ is the window structure built from $\Psi_j(q)$, and $\kappa$ is an indicator function to filter contextual facts based on zero or more criteria (\textit{i.e.,} entities, locations, or context).
The query window $\mathcal{W}_0(q)$ is constructed from only actors, recipients, and qualifiers of the query, and as such, it does not include relation information.

\subsection{Fact Representation}
For a given fact $\psi \in \Psi_i(q)$, we compute three role-specific DeepSet \citep{zaheer2017deep} representations for actors, recipients, and qualifiers as
{\small
\begin{equation}
\begin{split}
\mathbf{h}_{\psi}^A &= g_A\bigl(\text{Pool}(\{f_A(E_\mathcal{E}[e]) : e \in A_\psi\})\bigr),\\
\mathbf{h}_{\psi}^R &= g_R\bigl(\text{Pool}(\{f_R(E_\mathcal{E}[e]) : e \in R_\psi\})\bigr),\\
\mathbf{h}_{\psi}^Q &= g_Q\bigl(\text{Pool}\bigl(\bigl\{f_Q([E_\mathcal{R}[r];E_\mathcal{E}[e]]) : r,e \in Q_\psi\bigr\}\bigr)\bigr),
\end{split}
\end{equation}
}
where \textit{Pool} is a permutation invariant pooling function, $g_{\ast}$ and $f_{\ast}$ are MLPs with ReLU activations, and $E_\mathcal{E} \in \mathbb{R}^{|\mathcal{E}| \times d}$ and $E_\mathcal{R} \in \mathbb{R}^{|\mathcal{R}| \times d}$ are entity and relation embedding lookup tables, respectively.
While in the original DeepSet, \textit{Pool} is an elementwise mean pooling, we also experiment with max pooling, motivated by its utility in preserving important characteristics of input embeddings.
Given these role-specific representations, we construct the fact representation $z_{\psi}$ as
\begin{equation}
\begin{split}
    \tilde{z}_{\psi} & = f_\psi([h_{\psi}^A;h_{\psi}^R;h_{\psi}^Q]), \\
    (\alpha_{\psi}, \beta_{\psi}) & = f_{r}(E_\mathcal{R}[r_{\psi}]), \\
    z_{\psi} & = \alpha_{\psi} \odot \tilde{z}_{\psi} + \beta_{\psi},
\end{split}
\end{equation}
where $f_\psi$ and $f_r$ are MLPs, and $\alpha_{\psi}$ and $\beta_{\psi}$ impose a relation-specific affine transformation~\citep{perez2018film}.
For query facts, where we do not have access to relation information, we use a placeholder ``query" relation to avoid target leakage.

\begin{figure*}[t]
    \centering
    \includegraphics[width=\linewidth]{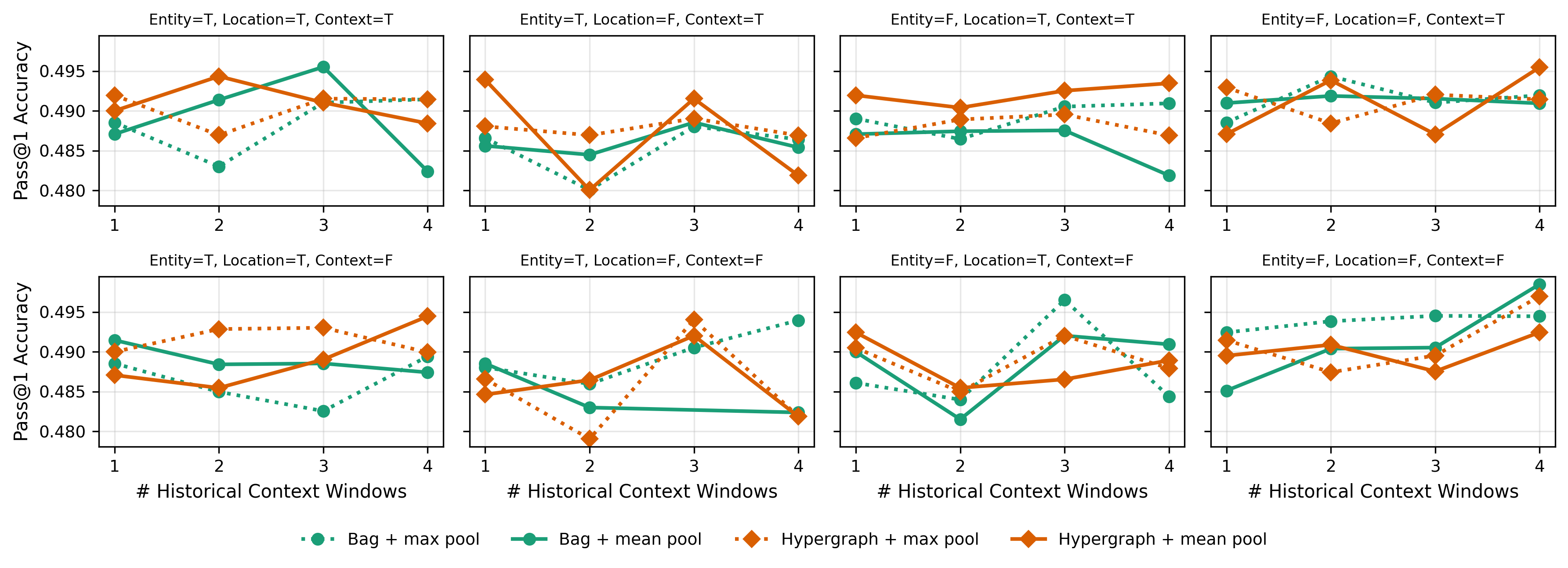}
    \caption{Relation prediction accuracy (\%) of GNN models on the final 24 months of the \texttt{htkgh-polecat} test set over the number of historical context windows used.}
    \label{fig:polecat_gnn}
\end{figure*}

\subsection{Window Representation}
Given a window $\mathcal{W}_i$ in $\mathcal{S}(q)$ with the corresponding fact set $\Psi_i(q)$, we compute a window representation as
\begin{equation}
z_i = f_\text{Pool}(\mathcal{W}_i) \in \mathbb{R}^d,
\end{equation}
where $d$ is the hidden dimension of the encoder.
Specifically, we consider two aggregation methods: \textit{bag aggregation} and \textit{hypergraph aggregation}.

\subsubsection{Bag Aggregation}
In this variation, we use a DeepSet aggregator to compute the window representation as
\begin{equation}
    z_i = g_\mathcal{W} (\text{Pool}(\{ f_\mathcal{W}(z_{\psi}): \psi \in \Psi_i(q) \})),
\end{equation}
where $g_\mathcal{W}$ and $f_\mathcal{W}$ are MLPs with ReLU activations.
This variation is an extreme simplification of the HTKGH structure as we naively aggregate facts within a window without any attention to their relationships.

\subsubsection{Hypergraph Aggregation}
To utilize relational information, we model each window as a typed hypergraph with facts for nodes and hyperedges between facts that share elements.
Specifically, for types $\tau \in \{A, R, Q\}$ we define three typed hyperedges as
\begin{equation}
    \mathcal{G}_i^{\tau}(e) = \{\psi \in \Psi_i(q) : e \in \tau_{\psi}\},
\end{equation}
where $\tau_{\psi}$ is the set of type-$\tau$ entities in $\psi$.

To perform hypergraph message passing, we first initialize node representations as
\begin{equation}
    x_{\psi}^{(0)} = z_{\psi} \  .
\end{equation}
Then, we compute a representation for each hyperedge by pooling from its nodes.
Specifically, the hyperedge representation at layer $l$ is computed as
\begin{equation}
\begin{split}
    & \bar{h}_{i,e}^{\tau,(l)} = \frac{1}{|\mathcal{G}_i^{\tau}(e)|}\sum_{\psi \in \mathcal{G}_i^{\tau}(e)} x_{\psi}^{(l)}, \\
    & h_{i,e}^{\tau,(l)} = f_e^{\tau}(\bar{h}_{i,e}^{\tau,(l)}),
\end{split}
\end{equation}
where $f_e^{\tau}$ is a type-specific MLP.
Finally, we use an aggregation scheme similar to R-GCN~\citep{schlichtkrull2018modeling} to update the node representation for $\psi \in \Psi_i(q)$ as
\begin{equation}
\begin{split}
    & m_{\psi}^{(l)} = \sum_{\tau \in \{A, R, Q\}} \sum_{e\in \tau_\psi} W_{\tau} h_{i,e}^{\tau, (l)}, \\
    & x_{\psi}^{(l+1)} = f_u([x_{\psi}^{(l)}; m_{\psi}^{(l)}]),
\end{split}
\end{equation}
where $W_{\tau}$ is a learnable matrix and $f_u$ is an MLP.

After $L$ layers of message passing, we use a DeepSet aggregator to compute the window representation as
\begin{equation}
\begin{split}
    z_i = g_\mathcal{W} (\text{Pool}(\{ f_\mathcal{W}(x_{\psi}^{(L)}): \psi \in \Psi_i(q) \})),
\end{split}
\end{equation}
where $g_\mathcal{W}$ and $f_\mathcal{W}$ are MLPs with ReLU activations.

\subsection{Temporal Encoder}
To aggregate information over all the windows and construct a final representation, we first create a context sequence as
\begin{equation}
    \mathcal{Z}(q) = [z_\mathcal{H}, ..., z_1, z_0] \in \mathbb{R}^{(\mathcal{H}+1) \times d} \  .
\end{equation}
Then, we compute the final representation as
\begin{equation}
h_q = f_t(\mathcal{Z}(q)) \in \mathbb{R}^d,
\end{equation}
where $f_t$ is a transformer~\citep{vaswani2017attention}.

\subsection{Linear Classifier}
Given the final representation $h_q$, we use a linear layer to perform relation classification as
\begin{equation}
\begin{split}
    p(. | \mathcal{S}(q)) = \text{Softmax}(W_{c} h_q + b)
\end{split}
\end{equation}
where $W_{c}$ and $b$ are learnable parameters.

\begin{figure}[t]
    \centering
    \includegraphics[width=\linewidth]{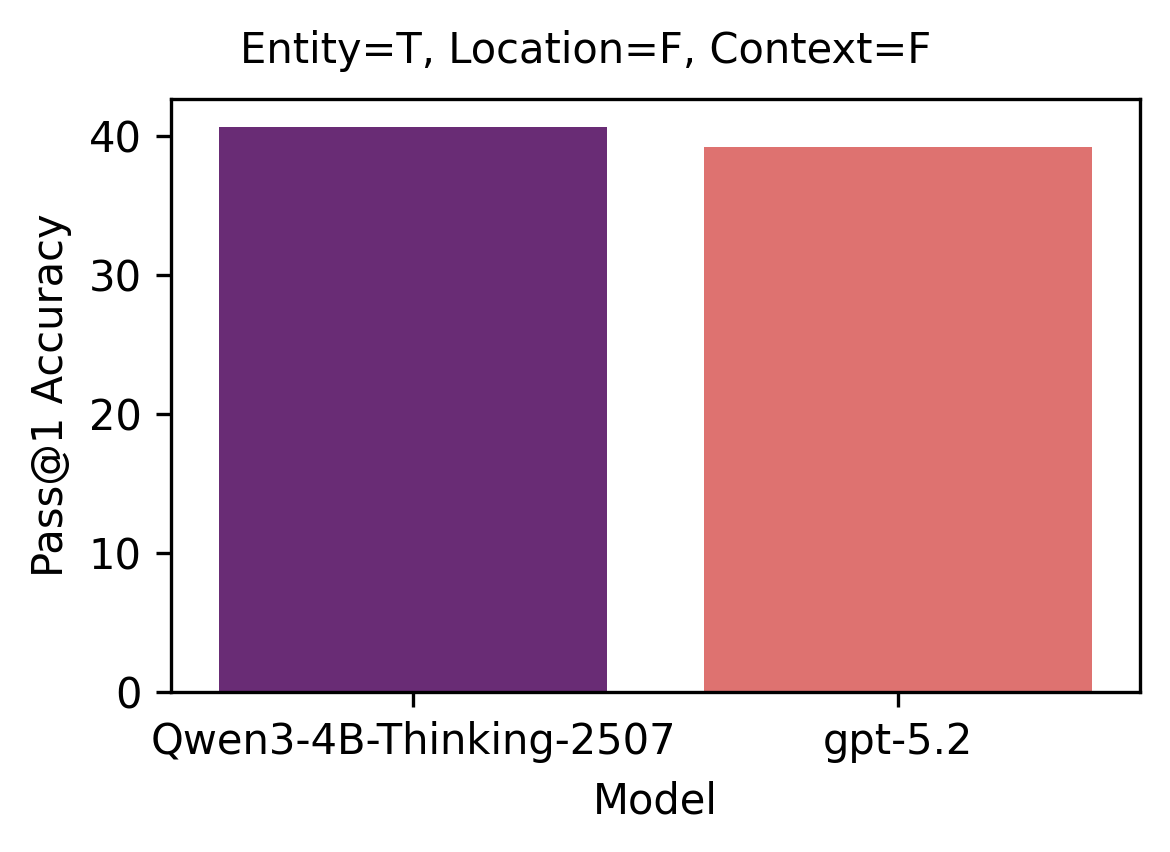}
    \caption{Relation prediction accuracy (\%) between the state-of-the-art open source and closed source thinking models.}
    \label{fig:gpt}
\end{figure}

\subsection{Loss Functions}
To train our models, we use the cross-entropy loss computed as
\begin{equation}
    \mathcal{L}_{\text{CE}} = -\frac{1}{N}\sum_{i=1}^N \text{log} ~ p(y_i | \mathcal{S}(q_i))
\end{equation}
where $y_i$ denotes the ground truth relation for query $i$ and $N$ is the batch size.

\subsection{Evaluation Setup}

\paragraph{Historical Context}
To construct the contextual information, we use a combination of the query-dependent entity, location, and context filters, analogously to those used in the LLM predictors.

\paragraph{Hyperparameters}
For each variant, we sweep over $learning\_rate \in \{3e^{-5}$, $1e^{-4}$, $3e^{-4}\}$ using AdamW optimizer with a weight decay of 0.01.
Training is done over eight epochs with a batch size of 256.
For the temporal encoder, we use a single-layer encoder-only Transformer with $d\_model=128$ and $nhead=4$.
For the hypergraph aggregator, we use $L=2$.
Finally, with each window covering one day, we temporally aggregate over 7 days before the query.

\subsection{Window Encoder Bottleneck Discussion}

While LLMs attend to individual facts in the context, our GNNs collapse all facts in a given window into one vector.
Based on our experimental results, this approach seems to be too coarse when the historical context is highly filtered.
Moreover, whether models use the bagging or hypergraph aggregation window encoder has little effect on performance.
Finally, we observe that using mean or max pooling in set encoders has little effect on the performance.

\subsection{Ablation on Number of Windows}
\autoref{fig:polecat_gnn} illustrates our experimental results when we vary the number of windows of context for our GNN-based models.
In these experiments, we do not observe any notable trends in model performance, which supports our information bottleneck hypothesis, where window representations contain a mix of signal and noise that is challenging for relation prediction.

\section{Closed Source vs. Open Source Models}
\autoref{fig:gpt} presents a comparison between the state-of-the-art open source and closed source thinking models.
Surprisingly, the open source model (\textit{i.e.,} \texttt{Qwen3-4B-Thinking-2507}) edges out the closed source model (\textit{i.e.,} \texttt{gpt-5.2}) by \textbf{\textcolor{PineGreen}{+1.4\%}}, showcasing an impressive performance on this task.